\documentclass[runningheads]{llncs}

\usepackage[mobile]{eccv}

\usepackage{eccvabbrv}

\usepackage{graphicx}
\usepackage{booktabs}
\usepackage{xcolor}
\usepackage{placeins}

\usepackage[accsupp]{axessibility} 
\usepackage{hyperref}
\usepackage{fontawesome5}

\begin{document}

\title{CAM3R: Camera-Agnostic Model for 3D Reconstruction} 

\author{Namitha Guruprasad\inst{1} \and
Abhay Yadav\inst{1} \and
Cheng Peng\inst{2} \and
Rama Chellappa \inst{1}}

\authorrunning{N.~Guruprasad et al.}

\institute{
Johns Hopkins University, Baltimore, MD 21218, USA\\
\email{\{ngurupr1, ayadav13, rchella4\}@jhu.edu}
\and
University of Virginia, Charlottesville, VA 22904, USA\\
\email{xuz7wn@virginia.edu}
}

\maketitle
\vspace{-15pt} 
\begin{center}
    \href{https://nam1410.github.io/cam3r}{\faCamera\ \textcolor{blue}{\footnotesize{Project Page}}}
\end{center}

\begin{figure}[ht]
  \centering
  \includegraphics[width=\linewidth,height=0.47\textheight]{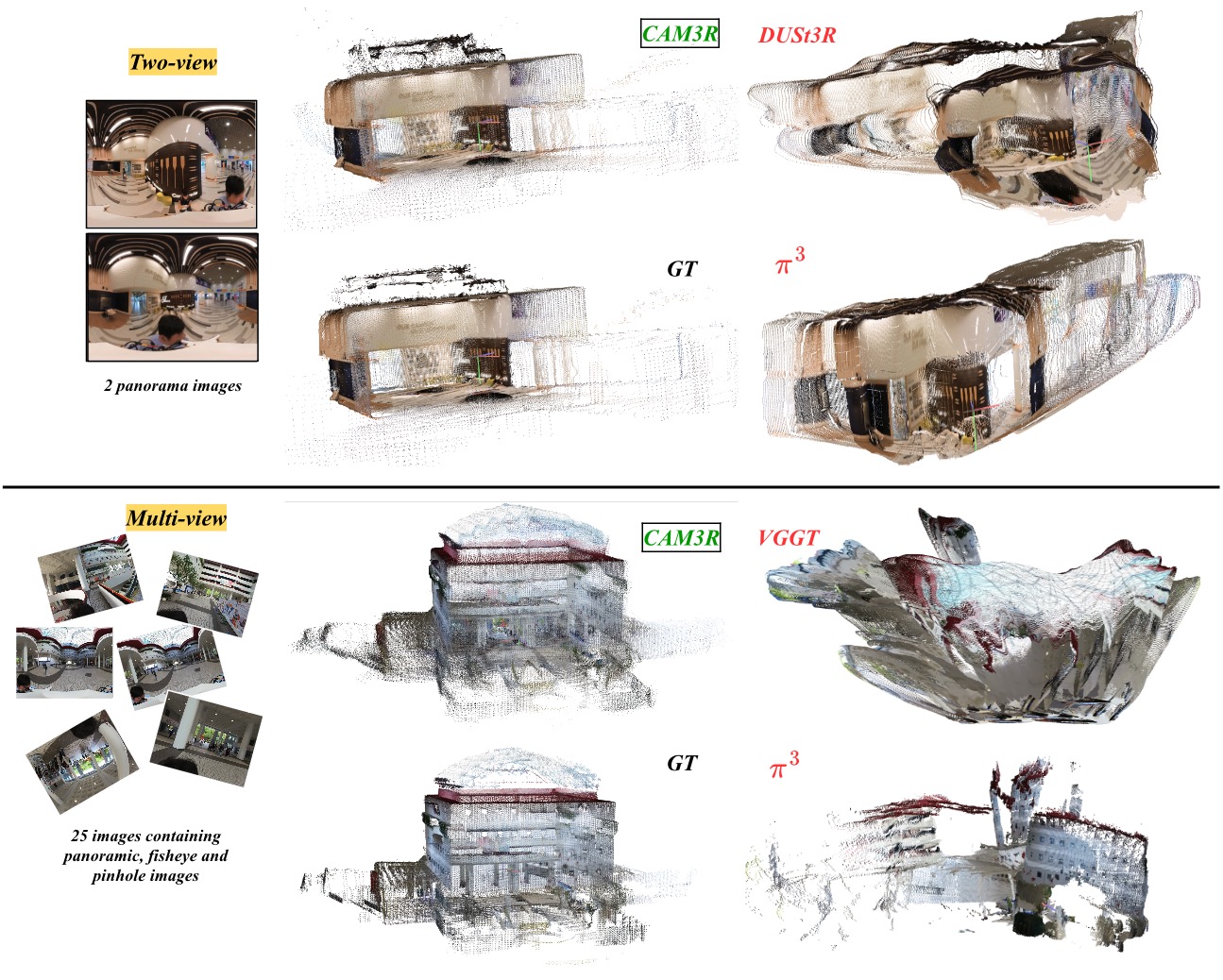}
  \caption{CAM3R provides a robust, feed-forward 3D reconstruction for two-view and multi-view scenarios across disparate optical manifolds, including pinhole, fisheye and panoramic cameras, where recent 3D foundation models fail. Above, we highlight CAM3R's performance on unseen scenes from the 360Loc dataset, visualizing both raw two-view predictions and multi-view reconstructions.}
  \label{fig:teaser}
  \vspace{-20pt} 
\end{figure}

\begin{abstract}
\sloppy
Recovering dense 3D geometry from unposed images remains a foundational challenge in computer vision. Current state-of-the-art models are predominantly trained on perspective datasets, which implicitly constrains them to a standard pinhole camera geometry. As a result, these models suffer from significant geometric degradation when applied to wide-angle imagery captured via non-rectilinear optics, such as fisheye or panoramic sensors. To address this, we present CAM3R, a Camera-Agnostic, feed-forward Model for 3D Reconstruction capable of processing images from wide-angle camera models without prior calibration. Our framework consists of a two-view network which is bifurcated into a Ray Module (RM) to estimate per-pixel ray directions and a Cross-view Module (CVM) to infer radial distance with confidence maps, pointmaps, and relative poses. To unify these pairwise predictions into a consistent 3D scene, we introduce a Ray-Aware Global Alignment framework for pose refinement and scale optimization while strictly preserving the predicted local geometry. Extensive experiments on various camera model datasets, including panorama, fisheye and pinhole imagery, demonstrate that CAM3R establishes a new state-of-the-art in pose estimation and reconstruction.
\end{abstract}

\section{Introduction}
\label{sec:intro}
The reconstruction of 3D environments from 2D imagery remains a cornerstone task in computer vision. Recent foundation models \cite{wang2024dust3r,leroy2024grounding,jang2025pow3r,wang2025vggt,wang2025pi} can achieve 3D reconstruction through a single feed-forward pass by learning a prior from large-scale 3D data. However, these models are grounded in the pinhole camera assumption because of its simple expression and availability. This bias leads to significant geometric artifacts when these models are evaluated on wide-angle imagery captured via non-linear optics, such as fisheye or $360^\circ$  panoramic lenses. 

To mitigate this, a common solution involves undistorting wide-angle images to simulate a rectified pinhole projection before processing via downstream reconstruction models. This rectification process forces extreme stretching or discards peripheral regions with high curvature \cite{sturm2011camera}. Furthermore, it often requires running Structure-from-Motion (SfM) or other optimizations, defeating the purpose of feed-forward models. 

Recent deep learning models directly establish the relationship between a 3D imaging ray and its corresponding pixel, making minimal assumptions regarding the underlying camera model and offering a more interpretable framework that aligns with human visual perception \cite{liao2025deeplearningcameracalibration}. However, these methods have been confined to monocular contexts \cite{piccinelli2025unik3d}, lacking a robust extension to multi-view settings. These observations prompt an investigation to check whether naive fine-tuning of such state-of-the-art models on wide-angle datasets is sufficient to rectify geometric degradation. As these models directly regress 3D structures in a single forward pass, they encounter optimization challenges when learning multi-view structures and non-linear projections arising from distorted camera geometry at the same time, which complicates the overall learning objective \cite{vandenhende2021multi, yu2020gradient}.

In this paper, we propose \textbf{CAM3R}, a \textbf{C}amera-\textbf{A}gnostic feed-forward \textbf{M}odel for \textbf{3}D \textbf{R}econstruction from uncalibrated and unposed imagery across a variety of camera types. The core of our architecture is a two-view network capable of regressing a dense scene representation without requiring prior intrinsic and extrinsic metadata. Our approach simultaneously addresses two challenges: (a) handling non-linear distortions in wide-angle imagery and (b) extracting 3D contextual cues for robust pose estimation. For the former, we propose a \textbf{Ray Module (RM)} designed for a two-view setting which learns a ray representation capable of processing wide-angle lens geometries, even when the input pair originates from heterogeneous camera models. For the latter, we introduce a \textbf{Cross-view Module (CVM)}, which learns the relationship between image pairs to generate 3D representations: radial distance with confidence maps, pointmaps, and the relative pose between the two views. To integrate pairwise predictions into a globally consistent 3D reconstruction, we introduce \textbf{Ray-Aware Global Alignment}. Traditional bundle adjustment often fails on distorted images due to the pinhole assumption, where pixel distances linearly correspond to 3D distances \cite{triggs1999bundle,wang2024dust3r}. Our Ray-Aware strategy lifts optimization into a 3D, ray-consistent space. Using constraints from the two-view network, we prune the scene graph to retain reliable edges and optimize camera poses and relative scales for global consistency.

In summary, our primary contributions are as follows:
\begin{itemize}
    \item \textbf{Unified Camera-Agnostic Framework:} We present \textbf{CAM3R}, a unified feed-forward pipeline that performs joint lens calibration and 3D reconstruction from unposed imagery across multiple camera geometries.
    \item \textbf{Decoupled Supervision Strategy:} We introduce a robust supervision strategy for the Ray Module and Cross-view Module across highly heterogeneous camera sources by jointly training rays, dense pointmaps, and relative poses.
    \item \textbf{Ray-Aware Global Alignment:} We propose a global alignment procedure that fuses local pairwise predictions into a globally consistent 3D coordinate system. Our method leverages a ray-aware optimization to maintain structural fidelity across diverse lens types.
    \item \textbf{State-of-The-Art Performance:} We demonstrate that our unified formulation achieves state-of-the-art results in two-view and multi-view reconstruction. Notably, CAM3R provides significant performance gains in challenging cross-modality scenarios, including large Field-of-View (FoV) and panoramic captures.
\end{itemize}

\section{Related Work}
\label{sec:relwork}
\subsection{Distorted Camera Intrinsics}
Camera intrinsic calibration involves estimating the parameters necessary to establish a mapping between the two-dimensional image plane and the three-dimensional coordinate system. Classical approaches \cite{brown1971close, conrady1919lens} rely on fixed parametric forms, which prove effective for narrow FoV sensors with negligible optical aberrations. However, they struggle with extreme FoV cameras. To support extreme FoV geometries, many models \cite{kannala2006generic, geyer2000unifying,urban2015improved} rely on higher-order parameters. However, they are prone to over-fitting due to complex parameter dependencies. Popular deep learning calibration frameworks\cite{bogdan2018deepcalib, veicht2024geocalib} infer focal length and distortion parameters directly from a single natural image. While these methods achieve reasonable results in canonical settings, they often exhibit limited generalization to unseen lens designs and are susceptible to training-set biases. In most downstream tasks, particularly 3D reconstruction, the fundamental objective is to determine the direction and distance of the light ray that formed each pixel \cite{liao2025deeplearningcameracalibration}. Building on this insight, recent works \cite{piccinelli2025unik3d} have shifted toward learning continuous representations that map each pixel to a 3D ray direction. However, these advancements remain largely confined to monocular settings. 

\subsection{Structure-from-Motion (SfM) and Multi-View Stereo (MVS)}
SfM formulates 3D reconstruction by simultaneously solving for sparse geometry and the underlying camera trajectory from an unordered image collection. Traditional frameworks, such as COLMAP \cite{schoenberger2016sfm}, operate through a rigid sequential pipeline: (a) extracting and matching local features \cite{lowe2004distinctive,detone2018superpoint,sarlin2020superglue,edstedt2024roma},  (b) solving two-view epipolar geometry (c) optimizing a multi-view sparse graph via global bundle adjustment \cite{triggs1999bundle}. To obtain dense surfaces, MVS \cite{schoenberger2016mvs} is employed to triangulate dense 3D geometry from SfM-calibrated viewpoints. However, such pipelines are highly susceptible to error propagation, where inaccuracies in initial SfM pose estimations degrade the fidelity of the downstream dense reconstruction. Recently, a paradigm shift has emerged with the introduction of geometric vision foundation models, such as DUSt3R \cite{wang2024dust3r}, MASt3R \cite{leroy2024grounding}, Pow3R \cite{jang2025pow3r}, VGGT \cite{wang2025vggt}, and $\pi^3$ \cite{wang2025pi}. These models bypass traditional multi-stage pipelines by directly regressing dense 3D pointmaps by implicitly capturing camera poses and scene geometry in a single feed-forward pass. Despite their versatility, these methods remain fundamentally limited by their reliance on the pinhole camera model, which dominates the training data and can lead to degraded performance on images from non-linear or wide-field-of-view lenses.

\section{Method}
In this section, we begin by providing a brief notation for a better understanding of CAM3R's architectural design. Following this, we describe our two-view feed-forward network comprising the Ray Module (RM) and the Cross-view Module (CVM), along with the multi-task supervised loss functions that govern their optimization. Finally, we introduce our Ray-Aware Global Alignment strategy to integrate the pairwise predictions from our two-view network into a globally consistent 3D reconstruction. 

\subsubsection{Notations.}
\begin{itemize}
    \item \textit{Input Data}: Consider a pair of uncalibrated views (images) $I_1 (\mathbf{u}), I_2 (\mathbf{u})\in \mathbb{R}^{W \times H \times 3}$ captured from varying lens geometries (panorama, fisheye, pinhole), where, $\mathbf{u} = (u,v)$ denotes the pixel coordinates for each image.
    \item \textit{Directional Rays:} Let $\mathbf{d}_i \in (\mathbb{S}^2)^{W \times H}$ represent the per-pixel directional rays for view $i$, where each unit vector $d_i(u,v) $ denotes the angular bearing of a pixel in the camera's local coordinate system. The mapping $\psi: \mathbb{R}^2 \mapsto \mathbb{S}^2$ projects 2D pixel coordinates $\mathbf{u}$ onto the spherical domain $(\theta, \phi)$. To represent these ray directions, we use Spherical Harmonic expansion where $Y_l^m$ denotes the basis functions of degree $l$ and order $m$. 
    \item \textit{Radial Distance and Confidence:} Let $r_i \in (\mathbb{R}^+)^{W \times H}$ represent the per-pixel Euclidean distance from the camera’s optical center along each corresponding directional ray $\mathbf{d}_i(\mathbf{u})$. To account for uncertainty, we predict a confidence map $\sigma_i \in (\mathbb{R^+})^{W \times H}$, where higher values indicate greater reliability in the estimated radial distance.
    \item \textit{Pointmap Representation:} Let $X^{i,i} \in \mathbb{R}^{W \times H \times 3}$ be the dense pointmap for a given view $i$, which is the element-wise product of the estimated ray directions and their corresponding radial distances:
    \begin{equation}
    X^{i,i}(\mathbf{u}) = \mathbf{d}_i(\mathbf{u}) \cdot r_i(\mathbf{u})
    \label{eq:rayDistDP}
    \end{equation}  
    \item \textit{Coordinate Transformations: }We denote $X^{i,j}$ as the pointmap $X^i$ expressed in the coordinate frame of camera $j$. Given camera poses $\mathbf{P}_i, \mathbf{P}_j \in SE(3)$, the transformation between frames is   $X^{i,j}(\mathbf{u}) =
    \mathbf{P}_j \mathbf{P}_i^{-1} X^{i,i}(\mathbf{u}).$
    
\end{itemize}

\subsection{CAM3R}
\begin{figure}[t]
  \centering
  \includegraphics[width=\linewidth]{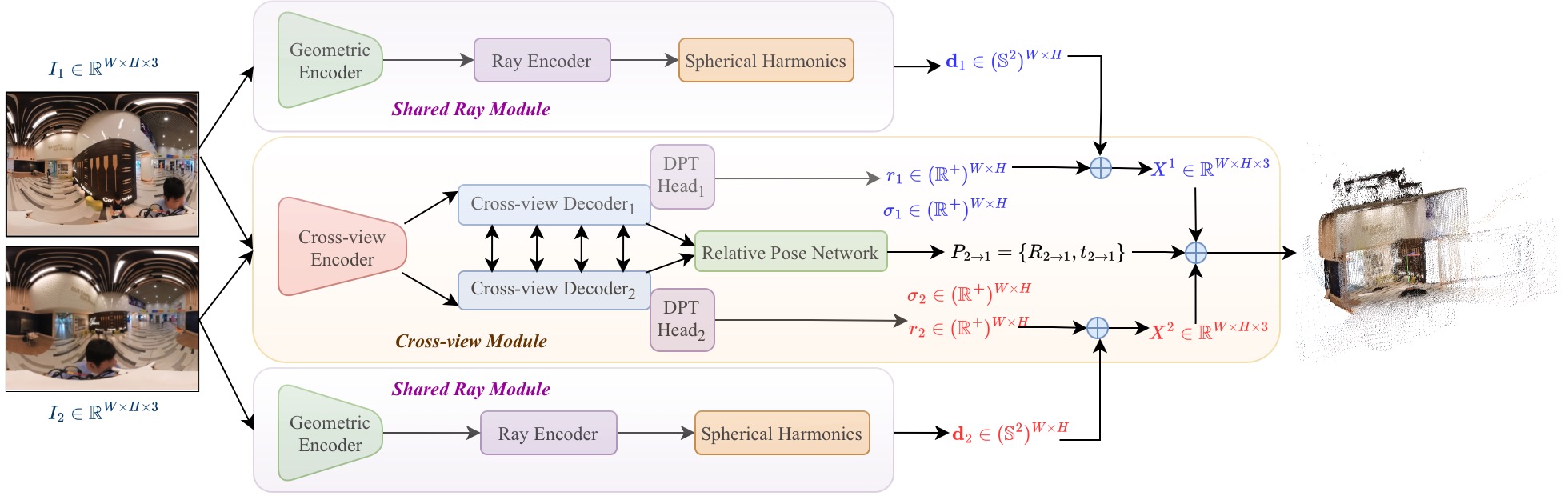}
  \caption{\textbf{CAM3R Overview.} Given an input image pair $(I_1, I_2)$, the framework operates through two parallel streams. The \textit{\textbf{Shared Ray Module}} recovers the internal camera geometry by regressing Spherical Harmonic coefficients to reconstruct continuous ray directional fields $\mathbf{d}_i$. Simultaneously, the \textbf{\textit{Cross-view Module}} extracts features and utilizes a dual-block transformer decoder to facilitate information exchange between the two views. Specialized DPT heads then regress radial distances $\mathbf{r}_i$ with confidence maps $\sigma_i$, while a Relative Pose Network estimates the rigid transformation $P_{2 \to 1}$. The local pointmaps $\mathbf{X}^{i,i}$ are generated by fusing rays $\mathbf{d}_i$ with radial distances $\mathbf{r}_i$. Finally, the second view is transformed into the reference coordinate frame of the first view via $P_{2 \to 1}$ to produce the globally aligned 3D reconstruction.}
  \label{fig:pipeline}
\end{figure}

We design CAM3R to address camera-agnostic 3D reconstruction via direct regression by drawing inspiration from UniK3D \cite{piccinelli2025unik3d} and DUSt3R \cite{wang2024dust3r}. A detailed rationale for this specific network topology is provided in the supplementary material (\cref{sec:arch}). As illustrated in \cref{fig:pipeline}, the network $\mathcal{F}$ consumes an image pair $(I_1, I_2)$ and jointly regresses per-pixel ray fields $\mathbf{d}_i$, radial distances $r_i$ with associated confidence $\sigma_i$, and the relative rigid pose $P_{2 \to 1}$. By decoupling local ray directions from radial distance, CAM3R provides greater flexibility in handling heterogeneous camera geometries. 

\subsubsection{Ray Module (RM)} serves as a backbone for estimating the internal camera geometry across pinhole, fisheye and panoramic lens models. For each input image $I_i$, a shared geometric encoder extracts class tokens $T_i$, which are then processed by a pair of Transformer Ray Encoder (T-Enc) layers. These layers output a compact set of SH expansion coefficients $\mathbf{c}_{l,m}^i$ and utilize a finite SH basis $Y_l^m$ up to a maximum degree $L$. This bandwidth parameter $L$ determines the high-frequency detail capturing capacity of the reconstructed camera manifold. A continuous per-pixel ray directional field $\mathbf{d}_i(\mathbf{u})$ is reconstructed via an inverse SH transformation. For any pixel coordinate  $\mathbf{u}$, the directional unit vector $\mathbf{d}_i(\mathbf{u})$ is:
\begin{equation}
\mathbf{d}_i(\mathbf{u}) =
\frac{
\sum_{l=1}^{L} \sum_{m=-l}^{l}
\mathbf{c}_{l,m}^{\,i} \, Y_l^m\!\left(\psi(\mathbf{u})\right)
}{
\left\|
\sum_{l=1}^{L} \sum_{m=-l}^{l}
\mathbf{c}_{l,m}^{\,i} \, Y_l^m\!\left(\psi(\mathbf{u})\right)
\right\|_2
}
\label{eq:RGM_SH}
\end{equation}

\subsubsection{Cross-view Module (CVM)} processes the input images $I_1, I_2$ through a Siamese Cross-view encoder to extract feature representations $F_1 = \text{Encoder}(I_1)$ and $F_2 = \text{Encoder}(I_2)$. These features are then passed to a dual transformer Cross-view decoder. Each decoder block $i \in \{1, \dots, B\}$ facilitates information exchange via self-attention and cross-attention mechanisms. Formally, the representations $G$ for each branch $v \in \{1, 2\}$ is updated as:
\begin{equation}
G_{1}^{i} = \mathrm{DecoderBlock}_{1}^{\,i}\!\left(G_{1}^{i-1},\, G_{2}^{i-1}\right),
\quad
G_{2}^{i} = \mathrm{DecoderBlock}_{2}^{\,i}\!\left(G_{2}^{i-1},\, G_{1}^{i-1}\right)
\label{eq:decoder_update}
\end{equation}
A Dense Prediction Transformer (DPT) head \cite{ranftl2021vision} is integrated with the decoder blocks to regress the radial distance $r_i$ and an associated confidence map $\sigma_i$ . We enforce strictly positive values $r_i(\mathbf{u}) \in \mathbb{R}^+$ via an activation layer. Using this radial distance with the per-pixel ray directions $\mathbf{d}_i(\mathbf{u})$ from the Ray Module, the 3D point coordinates are regressed  $X^{i,i}(\mathbf{u})$ as in \cref{eq:rayDistDP} in the local camera frame. To align the geometry of the second view  ($X^{2,2}$) with the reference frame of the first view ($X^{2,1}$), we employ a relative pose network attached to the decoder blocks. The relative pose network regresses a rigid transformation $P_{2\to1} = \{R_{2\to1}, \hat{\mathbf{t}}_{2\to1}\}$, where $R_{2\to1} \in SO(3)$ is the rotation matrix and $\hat{\mathbf{t}}_{2\to1} \in \mathbb{S}^2$ denotes a unit translation direction. The full translation vector is recovered by introducing a relative scale factor $s \in \mathbb{R}^+$ such that $\mathbf{t}_{2\to1} = s\,\hat{\mathbf{t}}_{2\to1}$. The transformation of the second pointmap $X^{2,2}$ into the coordinate system of the first view is then defined as

\begin{equation}
    X^{2,1}(\mathbf{u}) =
R_{2\to1}\,X^{2,2}(\mathbf{u}) + \mathbf{t}_{2\to1}.
\end{equation}

\subsection{Training Objectives}
\subsubsection{Asymmetric Angular Loss.}
To supervise the per-pixel directional rays $\mathbf{d}_i$ predicted by our model, we adopt an asymmetric angular loss inspired by UniK3D \cite{piccinelli2025unik3d}. Standard symmetric loss functions (\eg $L_2$, $L_1$) often bias neural networks toward the most frequent modes in the training data, which are typically narrow-FoV pinhole images. To prevent an inward collapse, we employ a quantile regression framework that enforces a significantly heavier penalty for angular underestimation than for overestimation. We first decompose the predicted dense ray field $\mathbf{d}_i$ into its corresponding spherical coordinates $(\hat{\theta},\hat{\phi}) $. Given the ground-truth angles $\theta^*$ and $\phi^*$, the loss for a specific target quantile $\alpha \in [0, 1]$ is formulated as:
\begin{equation}
\mathcal{L}_{\mathrm{AA}}^{\alpha}(\hat{\theta}, \theta^*)
=
\sum_{j \,:\, \hat{\theta}_j < \theta_j^*}
\alpha \, \left| \hat{\theta}_j - \theta_j^* \right|
+
\sum_{j \,:\, \hat{\theta}_j \ge \theta_j^*}
(1 - \alpha) \, \left| \hat{\theta}_j - \theta_j^* \right|
\label{eq:asym_angular}
\end{equation}
The total angular objective $\mathcal{L}_{A}$ is defined as a weighted combination of the losses for both spherical coordinates:
\begin{equation}
\mathcal{L}_{A}
=
\beta \, \mathcal{L}_{\mathrm{AA}}^{0.7}(\hat{\theta}, \theta^*)
+
(1-\beta) \, \mathcal{L}_{\mathrm{AA}}^{0.5}(\hat{\phi}, \phi^*)
\label{eq:total_angular}
\end{equation}

\subsubsection{Local Regression Loss.}
To supervise the predicted 3D pointmaps $X^{v,v}$, we minimize their distance to the ground-truth geometry in their respective local coordinate systems. This objective ensures that the model recovers independent 3D structures accurately, regardless of the relative camera poses between the two views. Given the ground-truth pointmaps $\overline{X}^{v,v}$ for view $v \in \{1, 2\}$, we adopt a Mean Squared Error (MSE) formulation. We denote $\mathcal{D}^v$ as the set of valid pixels with available ground truth. To resolve the scale ambiguity, we compute normalization factors for each view:
\begin{equation}
\eta_v
=
\operatorname{mean}_{\mathbf{u} \in \mathcal{D}^v}
\left\|
X^{v,v}(\mathbf{u})
\right\|_2,
\quad
\overline{\eta}_v
=
\operatorname{mean}_{\mathbf{u} \in \mathcal{D}^v}
\left\|
\overline{X}^{v,v}(\mathbf{u})
\right\|_2
\label{eq:local_scale_norm}
\end{equation}
where the normalization factor represents the average distance to the origin of all valid points in the respective frame. The point-wise regression loss $\mathcal{L}_{\mathrm{regr}}$ is then defined as:
\begin{equation}
\mathcal{L}_{\mathrm{regr}}
=
\sum_{v \in \{1,2\}}
\sum_{\mathbf{u} \in \mathcal{D}^v}
\left\|
\frac{1}{\eta_v} X^{v,v}(\mathbf{u})
-
\frac{1}{\overline{\eta}_v} \overline{X}^{v,v}(\mathbf{u})
\right\|_2^2
\label{eq:local_pointmap_loss}
\end{equation}

\subsubsection{Relative Pose Loss.}
To enforce multi-view consistency, we supervise the relative pose head by anchoring the translation magnitude to the scale of the reconstructed geometry. Let the ground-truth relative pose required to align the second view's pointmaps to the first view's coordinate frame be $\overline{P}_{2 \to 1} = \{\overline{R}_{2 \to 1}, \overline{\mathbf{t}}_{2 \to 1}\}$. Because our model regresses geometry up to an unknown scale, we resolve the translation ambiguity by computing scale factor $s$ from the predicted pointmaps. The scale factor is derived from the ratio of the predicted pointmap magnitudes to the ground-truth magnitudes and is detached from the gradient flow to serve as a static target. We then define the scale-adjusted ground-truth translation as $\mathbf{t}^*_{2 \to 1} = s \, \overline{\mathbf{t}}_{2 \to 1}$. The relative pose objective $\mathcal{L}_{pose}$ penalizes errors in both the rotation and the scale-anchored translation. For rotation, we utilize the geodesic distance on $SO(3)$, which provides a physically meaningful angular error. For translation, we utilize MSE loss to simultaneously supervise both the direction and the magnitude relative to the 3D scene:
\begin{equation}
\mathcal{L}_{\mathrm{rot}}
=
\arccos \left(
\frac{
\operatorname{Tr}
\!\left(
R_{2 \to 1}^{\top} \overline{R}_{2 \to 1}
\right)
- 1
}{2}
\right)
\label{eq:rot_loss}
\end{equation}
\begin{equation}
\mathcal{L}_{\mathrm{trans}}
=
\left\|
\mathbf{t}_{2 \to 1}
-
\mathbf{t}^{*}_{2 \to 1}
\right\|_2^2
\label{eq:trans_loss}
\end{equation}
The total pose objective is a weighted combination: $\mathcal{L}_{pose} = \lambda (\mathcal{L}_{rot} + \mathcal{L}_{trans})$. 

\subsubsection{Overall Optimization Objective.}
The final training objective for CAM3R is a joint optimization loss that unifies the geometric: Asymmetric Angular Loss ($\mathcal{L}_{A}$), structural: Local Regression Loss ($\mathcal{L}_{regr}$), and positional: Relative Pose Loss ($\mathcal{L}_{pose}$). The total loss $\mathcal{L}_{total}$ is a weighted summation of the above components:
\begin{equation}
\mathcal{L}_{\mathrm{total}}
=
\lambda_{A} \, \mathcal{L}_{A}
+
\lambda_{\mathrm{regr}} \, \mathcal{L}_{\mathrm{regr}}
+
\lambda_{\mathrm{pose}} \, \mathcal{L}_{\mathrm{pose}}
\label{eq:total_loss}
\end{equation}
This unified formulation allows the network to be trained end-to-end to learn the complex interplay between lens geometry and multi-view consistency of the 3D scene structure. 

\subsection{Ray-Aware Global Alignment}
\label{sec:RGA}
To reconstruct the entire scene from an unordered collection of images $\{I_1, \dots, I_N\}$, we propose a Ray-Aware Global Alignment framework. While conventional global alignment techniques like DUSt3R \cite{wang2024dust3r} implicitly or explicitly assume pinhole projection, our approach optimizes scene geometry within a purely ray-consistent 3D space. We perform the alignment through a multi-stage process.

\textbf{Scene-Graph Pruning.}
We construct an exhaustive graph $G(V, E)$, where edges $e = (i, j) \in E$ denote directed image pairs. Each edge undergoes a two-view inference pass with CAM3R, followed by a two-stage pruning protocol to filter inconsistent geometry: (a) \textbf{Symmetric Pose Consistency}: We evaluate bidirectional consistency between reciprocal edges $(i, j)$ and $(j, i)$. An edge is pruned if the angular deviation between the predicted rotation $R_{ji}$ and the transposed $R_{ij}^\top$ exceeds a threshold $\tau_{rot}$, or if the translation direction $\hat{t}_{ji}$ fails to align. This ensures that only mathematically reversible camera trajectories are retained for global optimization. (b) \textbf{Geometric Overlap Verification}: We compute strict Mutual Nearest Neighbor (MNN) correspondences in 3D space between $X^{i,i}$ and the transformed $X^{j,i}$. Edges with insufficient geometric overlap (\eg $<20\%$ of the pixel count) are discarded to prevent erroneous scene-graph linkages.

\textbf{Global Consensus and Optimization.}
Following pruning, we aggregate pairwise predictions into per-image consensus fields. For each image $I_i$, we compute the global ray field $\mathcal{D}_i$ by forming a confidence-weighted average of all normalized rays predicted across its valid incident edges. Given this canonical ray field, we initialize the global radial distance field $\mathcal{R}_i$ in three steps. First, we align pairwise radial distances along the consensus rays. Next, we resolve their relative scale using a robust median-based alignment. Finally, the aligned distances are fused via confidence-weighted averaging to produce the global field. The frozen consensus ray $\overline{\mathcal{D}_i}$ and radial $\overline{\mathcal{R}_i}$ fields define a per-pixel 3D point in the local coordinate system of camera $i$: $\mathbf{x}_i(\mathbf{u}) = \overline{\mathcal{R}_i}(\mathbf{u})\,\overline{\mathcal{D}_i}(\mathbf{u})$, which serve as geometric priors. The Global Alignment is formulated as an optimization over the set of camera poses  $\{P_i\}_{i=1}^N \in SE(3)$ and per-image scalar scales $$\{s_i\}_{i=1}^N \in \mathbb{R}^+.$$ Each scale $s_i$ compensates for the unknown global scale of the predicted radial distances for image $I_i$. Keeping the fixed geometric priors, let $\sigma_{i,j}(\mathbf{u})$ denote the predicted confidence for pixel $\mathbf{u}$ on edge $(i,j)$ from the pairwise model, which down-weights unreliable correspondences. Formally, the objective minimizes the confidence-weighted 3D alignment error over all connected pairs $(i,j)$ in the pruned scene graph $E$:
\begin{equation}
\min_{P_i, s_i}
\sum_{(i,j) \in E_{\text{pruned}}}
\sum_{\mathbf{u}}
\sigma_{i,j}(\mathbf{u})
\left\|
P_i \big( s_i \mathbf{x}_i(\mathbf{u}) \big)
-
P_j \big( s_j \mathbf{x}_j(\mathbf{u}) \big)
\right\|_2^2
\label{eq:global_opt_main}
\end{equation}
We employ a multi-stage alternating optimization scheme \cite{neal2011distributed} to obtain the global scene geometry. The process begins by fixing the scale factors to optimize the camera poses, followed by a scale correction step while the poses remain constant. This alternating cycle is repeated for a fixed number of iterations to stabilize the camera trajectory. Finally, we perform a joint optimization of both poses and scales to achieve global consistency across the entire scene graph. We provide additional details about Ray-Aware global alignment in the Supplementary (\cref{sec:RGAsupp}).

\section{Experiments}
\subsection{Training and Evaluation Dataset}
We train and test our framework on a collection of datasets encompassing panoramic, fisheye, and perspective imagery across indoor and outdoor environments. To ensure robust cross-model generalization, we augment existing datasets to induce optical heterogeneity:
\begin{itemize}
    \item 2D3DS \cite{armeni2017joint} and 360Loc \cite{huang2024360loc}: Originally panoramic collections, we synthesize corresponding fisheye and perspective views.
    \item ADT \cite{pan2023aria}: We expand this fisheye dataset by generating synthetic perspective counterparts.
    \item MegaDepth \cite{li2018megadepth}: Perspective dataset; no augmentation.
    \item CO3Dv2 \cite{reizenstein2021common}: This pinhole dataset is reserved for zero-shot evaluation and is not used during training. We expand this dataset by generating synthetic fisheye counterparts.
\end{itemize}

Throughout this paper, references to these datasets denote the above augmented, multi-modal versions. Data training and test splits mostly follow the official protocols established by the dataset authors unless otherwise specified. For datasets where image pairs are not explicitly defined, we curate them using ground-truth camera poses by ensuring significant spatial overlap and wide-baseline configurations. Specific details regarding the synthetic fisheye and pinhole projections and dataset splits are provided in the Supplementary Material (\cref{sec:Dat}). 

\subsection{Training Configuration}
\label{sec:trainingConfig}
We train our model using a two-phase curriculum \cite{bengio2009curriculum}. First, we optimize  our model exclusively on homogeneous pairs (\eg panorama-panorama) to establish a primary version \textbf{CAM3R-homo} for learning the intra-model multi-view geometry. In the second phase, we introduce heterogeneous pairs (\eg pinhole-panorama) to the training set to produce the final \textbf{CAM3R} model.  We use the AdamW optimizer \cite{loshchilov2019decoupledweightdecayregularization}  with an initial learning rate of $5 \times 10^{-5}$ with a linear warmup and cosine decay. Images are resized to $512$ pixels on the long edge. Training is conducted on four NVIDIA H200 GPUs using balanced dataset sampling to ensure stability across modalities. The ViT backbones \cite{dosovitskiy2021imageworth16x16words} of the  Ray Module and Cross-view Module are initialized with pre-trained weights of UniK3D \cite{piccinelli2025unik3d} and DUSt3R \cite{wang2024dust3r}, respectively. All other components in the network are initialized from scratch. Further specifications regarding the training curriculum are detailed in the Supplementary Material (\cref{subsec:training_config}). 

\subsection{Two-view Estimation}
\label{sec:twoview}
Relative pose estimation for a pair of images $I_1$ and $I_2$ involves recovering the rotation $R \in SO(3)$ and the translation direction $\mathbf{t} \in \mathbb{S}^2$ that maps the coordinate system of the second view into the first. In our unconstrained setting, this task is particularly challenging as the images may originate from disparate camera models (\eg a fisheye-panorama pair).

\textbf{Datasets.} We evaluate \textbf{CAM3R} on the held-out test splits of the augmented versions of 2D3DS, MegaDepth, ADT, and 360Loc, and further assess zero-shot generalization using the augmented CO3Dv2 dataset. Across all datasets, evaluation is performed on image pairs characterized by substantial spatial overlap and wide-baseline configurations, resulting in a challenging benchmark comprising both homogeneous and heterogeneous camera models. Comprehensive details regarding the pairwise sampling are provided in the Supplementary Material (\cref{sec:eval_protocols}).
\begin{figure}[th]
    \centering
    \includegraphics[width=\linewidth,height=0.46\textheight]{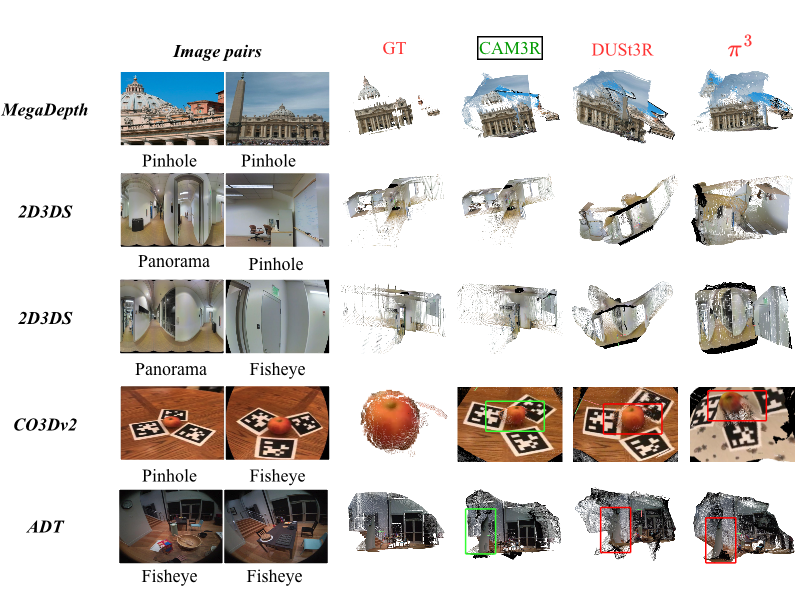}
    \caption{\textbf{Qualitative Two-View Reconstructions.} Visualization of 3D point clouds for image pairs across diverse optical manifolds (panorama, fisheye, pinhole). Despite extreme radial distortions and camera geometries, relative poses are accurately recovered and structural consistency is maintained. \textit{Note this is the raw output of the network.}}
    \label{fig:two_view_img}
\end{figure}

\textbf{Baselines and metrics.}  We compare CAM3R  against state-of-the-art foundation models: DUSt3R \cite{wang2024dust3r}, MASt3R \cite{leroy2024grounding}, Pow3R \cite{jang2025pow3r}, VGGT \cite{wang2025vggt}, and $\pi^3$ \cite{wang2025pi}. We report \textbf{Relative Rotation Accuracy} (RRA@15) and \textbf{Relative Translation Accuracy} (RTA@15), defined as the percentage of pairs with angular errors  $\tau$ below $ 15^\circ$.

\textbf{Results.}  As shown in Table \ref{tab:two_view_results}, CAM3R achieves superior performance overall. While $\pi^3$ and VGGT benefit from explicit training on MegaDepth and VGGT additionally on ADT, CAM3R remains highly competitive on unseen ADT scenes. Notably, on CO3Dv2, our model demonstrates strong zero-shot generalization by significantly outperforming all baselines in translation accuracy (88.2\% RTA@15). The most significant performance gap appears in cross-model scenarios, where DUSt3R and MASt3R collapse on panoramic datasets (360Loc, 2D3DS), CAM3R maintains a robust accuracy (\eg 97.7\% RRA@15; 94.3\% RTA@15 on 2D3DS). These quantitative gains are further evidenced by the qualitative point cloud reconstructions in \cref{fig:two_view_img}, which illustrate CAM3R’s ability to preserve structural integrity under extreme radial distortion.  In MegaDepth, CAM3R accurately recovers complex building architectures such as domes and tomb structures. Crucially, in fisheye sequences from ADT, CAM3R (marked \textcolor{green}{green} in \cref{fig:two_view_img}) preserves wall planes and clocks, which appear significantly distorted (\textit{bent}) in baseline (marked \textcolor{red}{red} in \cref{fig:two_view_img}) reconstructions. 

\begin{table}
\centering
\caption{Two-view relative pose estimation results. We report Accuracy at $15^\circ$ (higher is better). Bold indicates best performance.}
\label{tab:two_view_results}
\resizebox{\textwidth}{!}{
\begin{tabular}{@{}lcccccccccc@{}}
\toprule
\textbf{Model} & \multicolumn{2}{c}{\textbf{2D3DS}} & \multicolumn{2}{c}{\textbf{MegaDepth}} & \multicolumn{2}{c}{\textbf{CO3Dv2 (Zero-shot)}} & \multicolumn{2}{c}{\textbf{360Loc}} & \multicolumn{2}{c}{\textbf{ADT}} \\
\cmidrule(lr){2-3} \cmidrule(lr){4-5} \cmidrule(lr){6-7} \cmidrule(lr){8-9} \cmidrule(lr){10-11}
& RRA@15 & RTA@15 & RRA@15 & RTA@15 & RRA@15 & RTA@15 & RRA@15 & RTA@15 & RRA@15 & RTA@15 \\
\midrule
\textbf{DUSt3R} \cite{wang2024dust3r} & 10.6 & 6.0 & 95.6 & 80.8 & 94.7 & 43.1 & 0.0 & 0.0 & 91.0 & 63.6 \\
\textbf{MASt3R} \cite{leroy2024grounding} & 18.3 & 9.3 & 69.7 & 56.4 & \textbf{98.4} & 33.4 & 39.8 & 5.3 & 96.6 & 63.5 \\
\textbf{Pow3R} \cite{jang2025pow3r} & 7.5 & 6.0 & 96.2 & 74.2 & 95.8 & 38.3 & 0.0 & 0.0 & 96.6 & 79.2 \\
\textbf{VGGT} \cite{wang2025vggt} & 11.8 & 11.0 & 98.0 & 88.2 & 90.9 & 29.4 & 37.8 & 11.1 & 92.7 & 82.9 \\
\textbf{$\pi^3$} \cite{wang2025pi} & 16.8 & 11.4 & \textbf{99.8} & 93.3 & 90.7 & 22.7 & 38.5 & 13.0 & 97.5 & 93.8 \\
\midrule
\textbf{CAM3R-homo} & 65.4 & 56.8 & 97.2 & 92.6 & 96.1 & 66.5 & 58.3 & 54.7 & 98.2 & 93.4 \\
\textbf{CAM3R} & \textbf{97.7} & \textbf{94.3} & 96.8 & \textbf{94.2} & 97.5 & \textbf{88.2} & \textbf{96.0} & \textbf{91.0} & \textbf{99.0} & \textbf{95.0} \\
\bottomrule
\end{tabular}}
\end{table}

\subsection{Multi-view Reconstruction}
\begin{figure}
    \centering
    \includegraphics[width=0.8\linewidth]{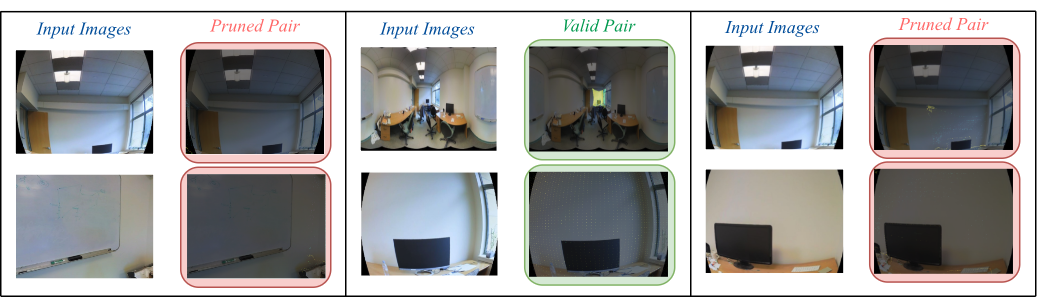}
    \caption{\textbf{Qualitative Pruning Analysis.} From left to right: (1) Successful rejection of a non-overlapping pair; (2) A valid pair with dense 3D correspondences; (3) Rejection of a doppelganger case where visually similar computer monitors yield inconsistent relative geometry.}
    \label{fig:pruning_viz}
\end{figure}

Multi-view relative pose estimation extends the two-view problem to a set of $N > 2$ images, where the goal is to recover a globally consistent set of camera poses $\{P_1, \dots, P_N\}$ in a unified coordinate frame. 
To recover the global scene, we first perform pairwise inference with \textbf{CAM3R} across the exhaustive scene graph. We then apply our two-stage pruning (\cref{sec:RGA}) to filter inconsistent edges as illustrated in \cref{fig:pruning_viz} and utilize Ray-Aware Global Alignment to resolve the final camera poses.

\textbf{Dataset.} Similar to two-view evaluation, we assess the multi-view performance of \textbf{CAM3R} on 2D3DS, MegaDepth, 360Loc, ADT, and CO3Dv2 (for zero-shot evaluation). Unlike the two-view evaluation in \cref{sec:twoview}, which relies on curated image pairs, the multi-view benchmark utilizes unstructured pools of images sampled from within a scene or subscene. Further technical specifics regarding the group-wise sampling protocols are detailed in the Supplementary Material (\cref{sec:eval_protocols}) .

\textbf{Baselines and Metrics.}  Due to the performance degradation of DUSt3R, MASt3R, and Pow3R on heterogeneous optics in the two-view setting, we limit our multi-view comparative analysis to VGGT and $\pi^3$. We report \textbf{Accuracy (RRA@30/RTA@30)}, mean \textbf{Average Accuracy (mAA@30)}, and \textbf{Absolute Trajectory Error (ATE)}. ATE is computed via root-mean-square error (RMSE) after Umeyama alignment \cite{kabsch1976solution} to account for global scale and pose.
\begin{figure}
    \centering
    \includegraphics[width=\linewidth,height=0.30\textheight]{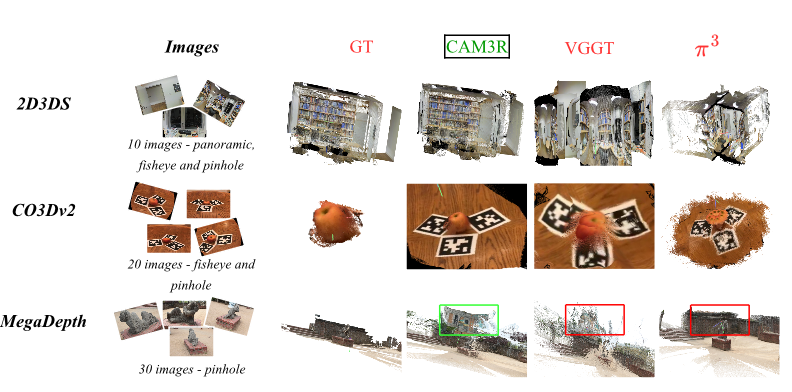}
    \caption{\textbf{Qualitative Multi-View Reconstructions.} Global camera trajectories and dense point clouds recovered from unstructured image pools across diverse datasets. Despite high radial distortion and lack of scenegraph information, globally consistent poses and structural geometry are maintained through the Ray-Aware Global Alignment, effectively mitigating trajectory drift and scale ambiguity.}
    \label{fig:multi_view_img}
\end{figure}

\textbf{Results.}  As shown in \cref{tab:multi_view_acc}, CAM3R achieves the highest average performance across the majority of datasets. VGGT and $\pi^3$ struggle with 2D3DS and 360Loc datasets, whereas CAM3R demonstrates robustness to these wide-angle geometries, achieving 73.5\% and 82.6\% mAA, respectively. While $\pi^3$ shows competitive accuracy on standard perspective dataset like MegaDepth, it benefits from explicit training on that domain. On the CO3Dv2 benchmark, CAM3R achieves a robust 64.9\% mAA in a zero-shot setting, while baseline models struggle to handle the mixture of fisheye and pinhole images. In \cref{tab:multi_view_ate}, CAM3R significantly reduces drift in wide-angle scenarios, reaching an ATE of $2.7$ on 360Loc compared to the baselines. These quantitative gains are reflected in the qualitative multi-view reconstructions in \cref{fig:multi_view_img}, which illustrate our model’s ability to recover globally consistent camera trajectories and dense scene geometry from unstructured, distorted image pools. In CO3Dv2, CAM3R (marked \textcolor{green}{green})  recovers well-defined object geometries, (here, the apple structure), whereas baseline models (marked \textcolor{red}{red}) produce diffused or scattered pointclouds. Furthermore, in MegaDepth, our model achieves superior structural completion of complex buildings; while some background noise is present due to sky regions, the primary architectural integrity is preserved more effectively than in baselines.

\begin{table}
\centering
\caption{Multi-view relative pose estimation results. We report Accuracy at $30^\circ$ (RRA and RTA) and mAA@30 (higher is better). Bold indicates best performance.}
\label{tab:multi_view_acc}
\resizebox{\textwidth}{!}{
\begin{tabular}{@{}lccccccccccccccc@{}}
\toprule
\textbf{Model} & \multicolumn{3}{c}{\textbf{2D3DS}} & \multicolumn{3}{c}{\textbf{MegaDepth}} & \multicolumn{3}{c}{\textbf{CO3Dv2(Zero-Shot)}} & \multicolumn{3}{c}{\textbf{360Loc}} & \multicolumn{3}{c}{\textbf{ADT}} \\
\cmidrule(lr){2-4} \cmidrule(lr){5-7} \cmidrule(lr){8-10} \cmidrule(lr){11-13} \cmidrule(lr){14-16}
& \textbf{RRA} & \textbf{RTA} & \textbf{mAA} & \textbf{RRA} & \textbf{RTA} & \textbf{mAA} & \textbf{RRA} & \textbf{RTA} & \textbf{mAA} & \textbf{RRA} & \textbf{RTA} & \textbf{mAA} & \textbf{RRA} & \textbf{RTA} & \textbf{mAA} \\
\midrule
\textbf{VGGT} & 31.8 & 34.4 & 7.6 & \textbf{100.0} & 97.4 & 68.8 & 70.5 & 75.3 & 19.6 & 47.9 & 50.8 & 19.5 & \textbf{100.0} & 95.6 & 60.3 \\
\textbf{$\pi^3$} & 40.0 & 35.8 & 9.6 & \textbf{100.0} & \textbf{98.4} & 73.4 & \textbf{89.5} & \textbf{91.6} & 22.7 & 48.6 & 47.4 & 17.8 & \textbf{100.0} & \textbf{100.0} & 75.8 \\
\midrule
\textbf{CAM3R + DUSt3R GA} & 72.3 & 55.1 & 38.8 & 86.5 & 72.3 & 68.5 & 69.7 & 56.3 & 46.1 & 66.7 & 59.2 & 55.5 & 78.3 & 71.1 & 70.8 \\
\textbf{CAM3R + Our GA} & \textbf{94.0} & \textbf{91.5} & \textbf{73.5} & 96.6 & 96.3 & \textbf{87.4} & 85.0 & 85.2 & \textbf{64.9} & \textbf{98.7} & \textbf{91.2} & \textbf{82.6} & 92.2 & 91.5 & \textbf{77.3} \\
\bottomrule
\end{tabular}}
\end{table}

\begin{table}
\centering
\caption{Multi-view trajectory alignment error (ATE RMSE). Lower is better. Bold indicates best performance.}
\label{tab:multi_view_ate}
\resizebox{0.7\textwidth}{!}{
\begin{tabular}{@{}lccccc@{}}
\toprule
\textbf{Model} & \textbf{2D3DS} & \textbf{MegaDepth} & \textbf{CO3Dv2(Zero-shot)} & \textbf{360Loc} & \textbf{ADT} \\
\midrule
\textbf{VGGT} & 3.8 & 0.7 & 1.3 & 6.3 & 0.5 \\
\textbf{$\pi^3$} & 2.9 & \textbf{0.6} & \textbf{0.7} & 5.8 & \textbf{0.4} \\
\midrule
\textbf{CAM3R + DUSt3R BA} & 2.4 & 1.2 & 1.6 & 4.5 & 0.6 \\
\textbf{CAM3R + Our BA} & \textbf{1.8} & 0.8 & 1.1 & \textbf{2.7} &\textbf{0.4} \\
\bottomrule
\end{tabular}}
\end{table}

\subsection{Ablation Studies}
\textbf{Impact of Heterogeneous Training.} We evaluate the necessity of the second training phase by comparing the full \textbf{CAM3R} against the phase-one baseline, \textbf{CAM3R-homo}. As shown in \cref{tab:two_view_results},  the inclusion of heterogeneous pairs (\eg panorama-perspective) during the second stage of our training curriculum significantly improves generalization for two-view relative pose estimation.

\textbf{Ray-Aware Global Alignment.} We evaluate the efficacy of our proposed Ray-Aware Global Alignment against the global alignment utilized in DUSt3R. As shown in \cref{tab:multi_view_acc}, our ray-based optimizer when applied to the pairwise predictions of CAM3R, demonstrates superiority across all datasets, particularly on heterogeneous datasets where mAA falls by nearly 35\%. Furthermore, the ATE RMSE results in \cref{tab:multi_view_ate} confirm that our approach effectively suppresses cumulative drift, reducing trajectory error by up to 40\% (\eg from 4.5 to 2.7 on 360Loc). 

\section{Conclusion}
In this paper, we address two challenges in 3D reconstruction: handling non-linear distortions in wide-angle imagery and extracting reliable cross-view cues for robust pose estimation. To this end, we introduce CAM3R, a regression-based framework designed to tackle both problems. The Ray Module learns pixel-wise ray representations for wide-angle lenses. The Cross-view Module models relationships between image pairs to predict 3D representations such as pointmaps and relative poses. Building on these components, we extend the framework to multi-view reconstruction of unordered, distorted images through a Ray-Aware Global Alignment method. Extensive evaluations demonstrate that CAM3R achieves state-of-the-art performance across pinhole, fisheye, and panoramic imagery, and successfully recovers globally consistent camera trajectories in challenging multi-view scenarios where traditional baselines fail to converge. In future work, we aim to unify the geometric and cross-view encoders into a more efficient backbone to enable real-time applications. We also plan to explore advanced ray-based positional encoding strategies to better capture high-frequency scene details, and to develop more sophisticated scene-graph pruning mechanism that can handle visual ambiguities like extreme viewpoint and appearance changes.

\section*{Acknowledgements}
We thank Anand Bhattad for helpful discussions and valuable feedback.
This research is based upon work supported by the Office of the Director of National Intelligence (ODNI), Intelligence Advanced Research Projects Activity (IARPA), via IARPA R\&D Contract No. 140D0423C0076. The views and conclusions contained herein are those of the authors and should not be interpreted as necessarily representing the official policies or endorsements, either expressed or implied, of the ODNI, IARPA, or the U.S. Government. The U.S. Government is authorized to reproduce and distribute reprints for Governmental purposes notwithstanding any copyright annotation thereon.

\clearpage

\setcounter{section}{0}
\setcounter{figure}{0}
\setcounter{table}{0}
\setcounter{equation}{0}

\renewcommand{\thesection}{\Alph{section}}
\renewcommand{\thefigure}{S\arabic{figure}}
\renewcommand{\thetable}{S\arabic{table}}
\renewcommand{\theequation}{S\arabic{equation}}

\renewcommand{\theHsection}{\Alph{section}}
\renewcommand{\theHfigure}{S\arabic{figure}}
\renewcommand{\theHtable}{S\arabic{table}}
\renewcommand{\theHequation}{S\arabic{equation}}

\vspace*{1cm}
\begin{center}
    \LARGE \bfseries Supplementary Material
\end{center}
\vspace{1cm}

\begin{figure}[ht]
    \centering
    \includegraphics[width=\linewidth,height=0.5\textheight]{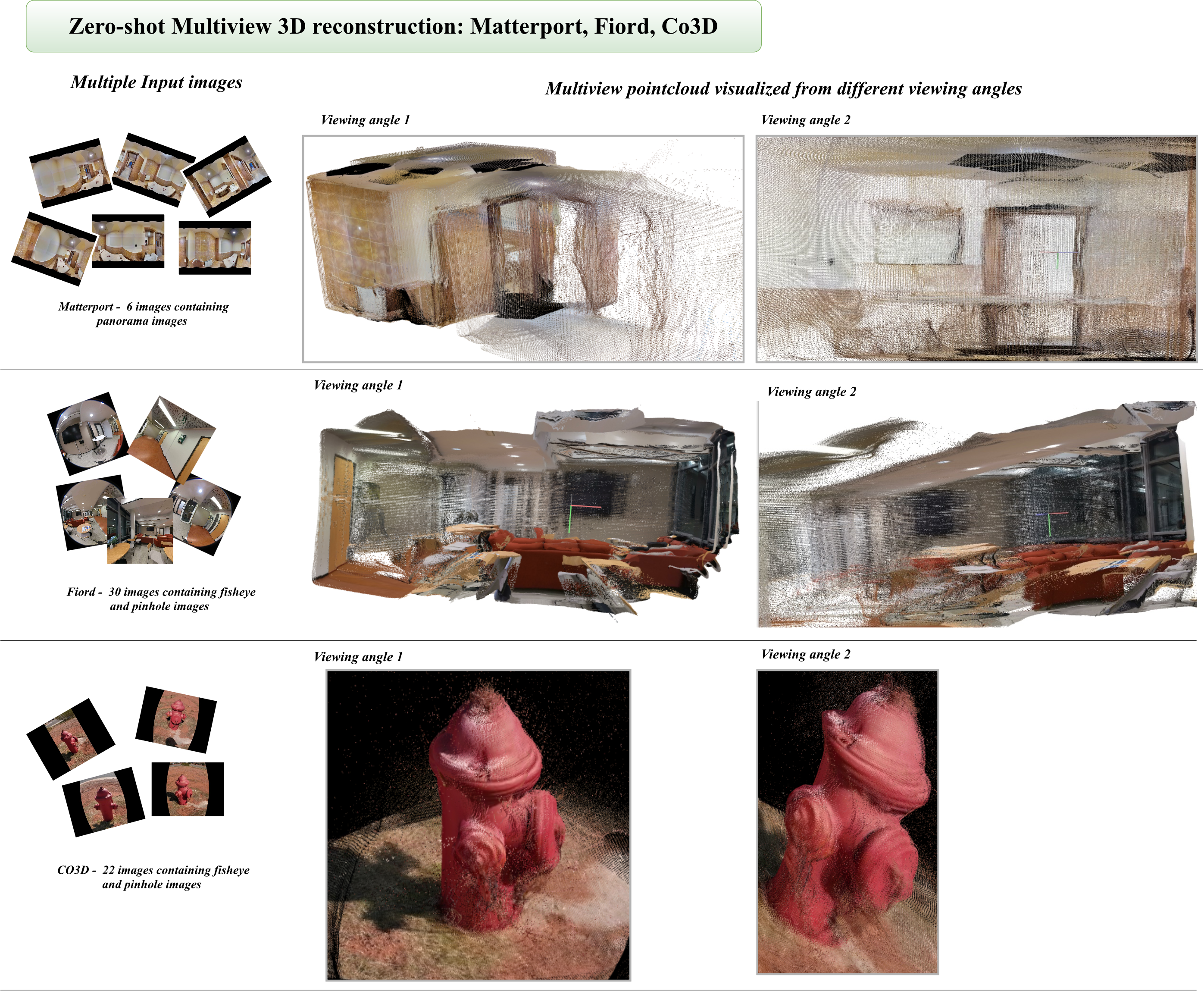}
    \caption{\textbf{Zero-shot generalization to unseen datasets.} CAM3R successfully recovers dense 3D point clouds on Matterport \cite{chang2017matterport3d}, FIORD \cite{gunes2025fiord}, and CO3Dv2 \cite{reizenstein2021common} despite no multi-view training on these domains. While the Ray Module leverages prior exposure to Matterport's optical manifold via UniK3D \cite{piccinelli2025unik3d} initialization, the Cross-View Module demonstrates true zero-shot structural generalization. Notably, in the FIORD dataset, CAM3R unwraps extreme 2D fisheye aberrations, strictly preserving rectilinear structures (\eg straight walls and sharp corners) in 3D space. To simulate unconstrained real-world environments, these scenes contain a heterogeneous mix of panoramic, fisheye, and pinhole captures. Multiple viewing angles are provided to highlight the structural integrity of the reconstructions.}
    \label{fig:multi_view_zeroshot}
    \vspace{-20pt}
\end{figure}

\begin{figure}[!t]
    \centering
    \includegraphics[width=\linewidth,height=0.5\textheight]{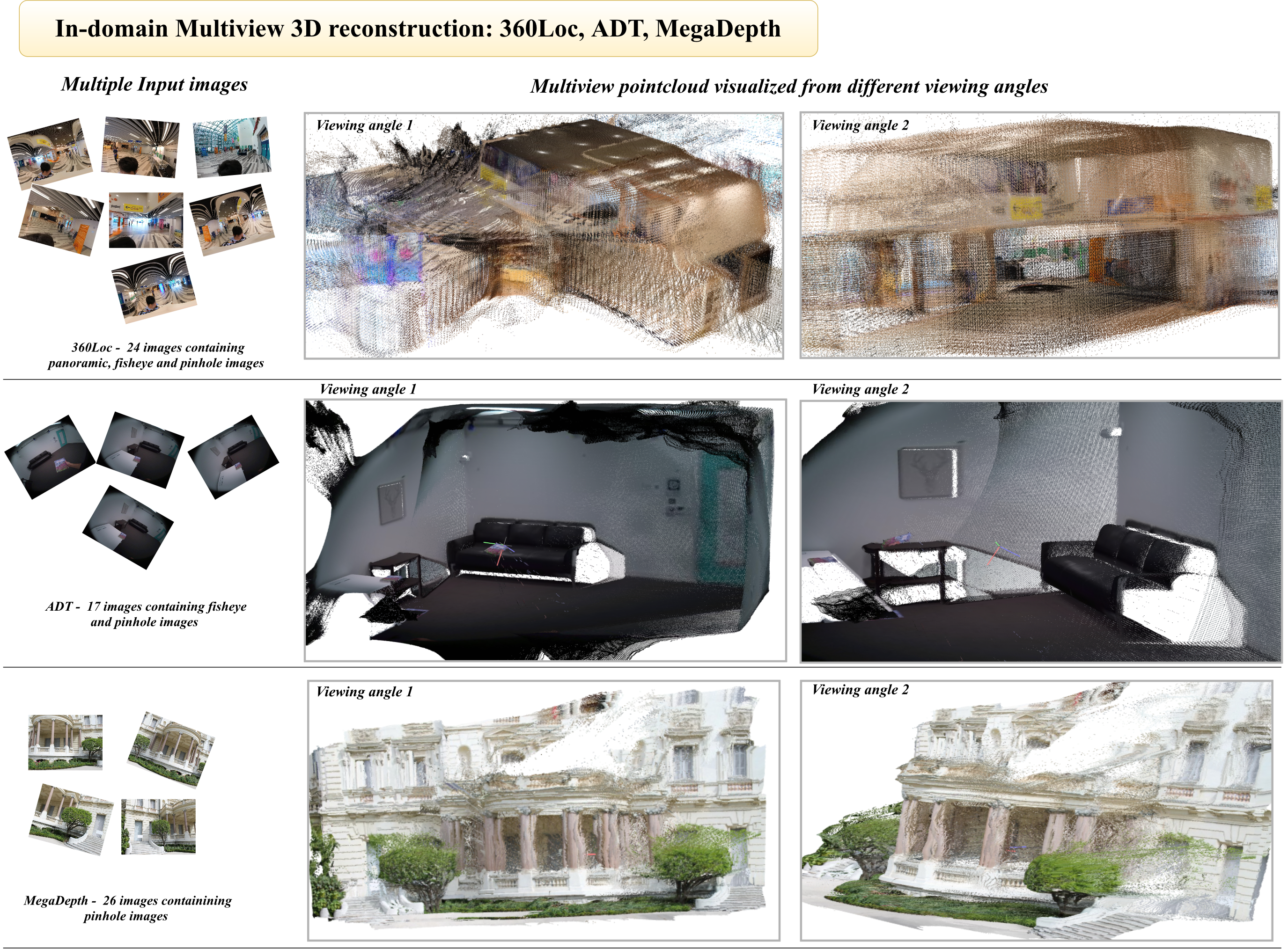}
    \caption{\textbf{In-domain multi-view 3D reconstructions.} Extended qualitative results from the 360Loc \cite{huang2024360loc}, ADT \cite{pan2023aria}, and MegaDepth \cite{li2018megadepth} test splits demonstrate CAM3R's robustness. In 360Loc, the model reconstructs expansive concourse regions despite the prevalence of smooth, highly reflective, and textureless surfaces that typically break traditional matching heuristics. For ADT, which features enclosed room scenes with dense, short-baseline egocentric fisheye captures, our rigorous scene-graph pruning yield a coherent pointcloud. Multiple viewing angles are provided to illustrate the dense coverage and structural consistency of the fused multi-modal outputs.}
    \label{fig:multi_view_indomain}
    \vspace{-20pt}
\end{figure}

\begin{figure}[!t]
    \centering
    \includegraphics[width=\linewidth,height=0.75\textheight]{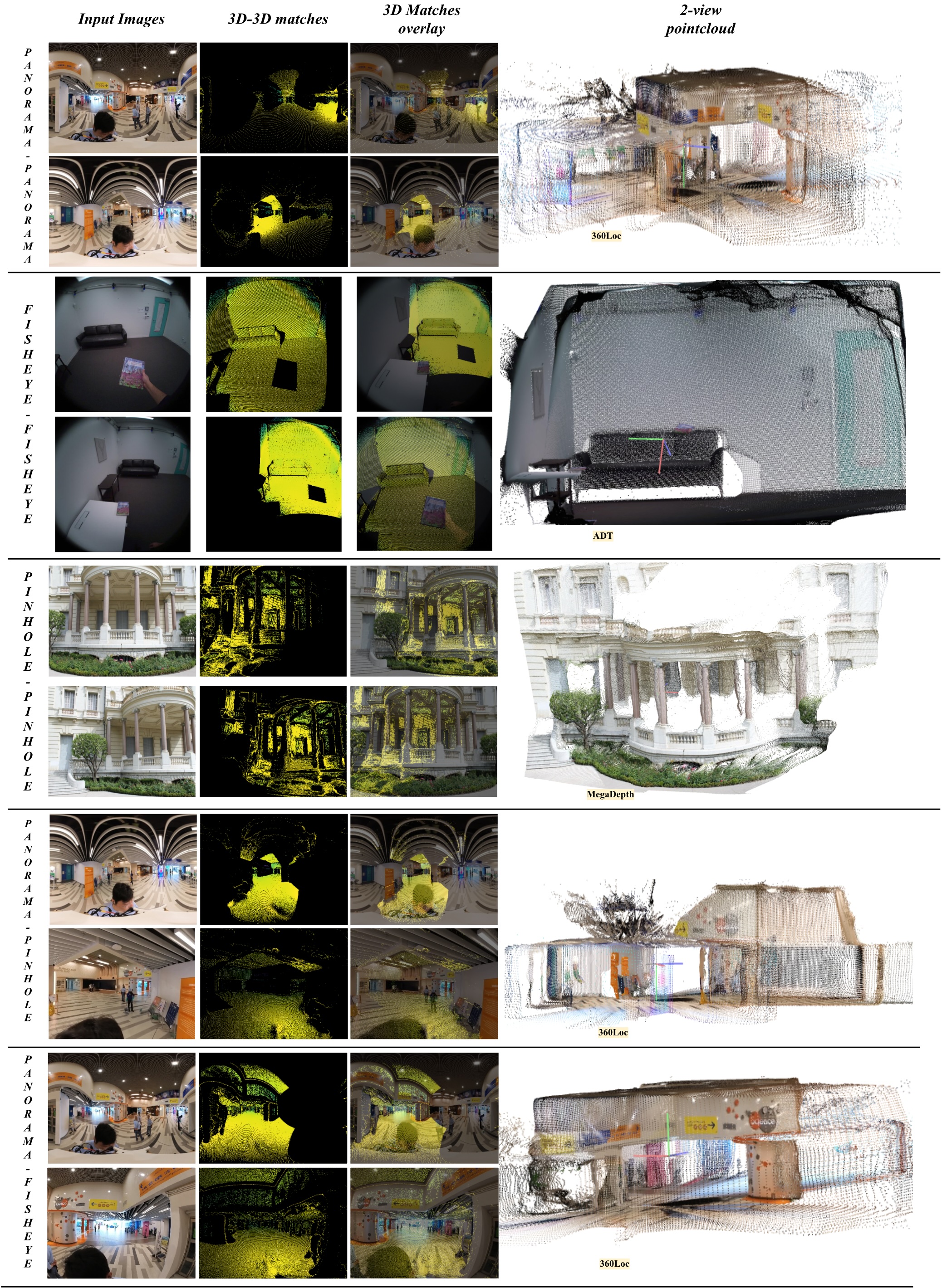}
    \caption{\textbf{Raw two-view reconstructions and cross-modal correspondences.} We visualize constituent image pairs from the multi-view scenes evaluated in \cref{fig:multi_view_indomain}. CAM3R demonstrates robustness in recovering two-view relative geometry not only for homogeneous pairs (panorama-panorama, fisheye-fisheye, pinhole-pinhole) but crucially across highly heterogeneous projection models (\eg panorama-pinhole and panorama-fisheye in 360Loc). We overlay 3D-3D Mutual Nearest Neighbor (MNN) matches utilized during our graph pruning phase, color-coded by the network's predicted distance (brighter yellow indicates stronger matches and closer points in 3D). \textit{Note: These visualizations represent the raw, feed-forward output of the two-view network prior to Ray-Aware Global Alignment.}}
    \label{fig:two_view_img_supp}
\end{figure}

\FloatBarrier

\section{Architecture Rationale: The Need for Decoupling}
\label{sec:arch}
\begin{figure*}[ht]
    \centering
    \includegraphics[width=\textwidth]{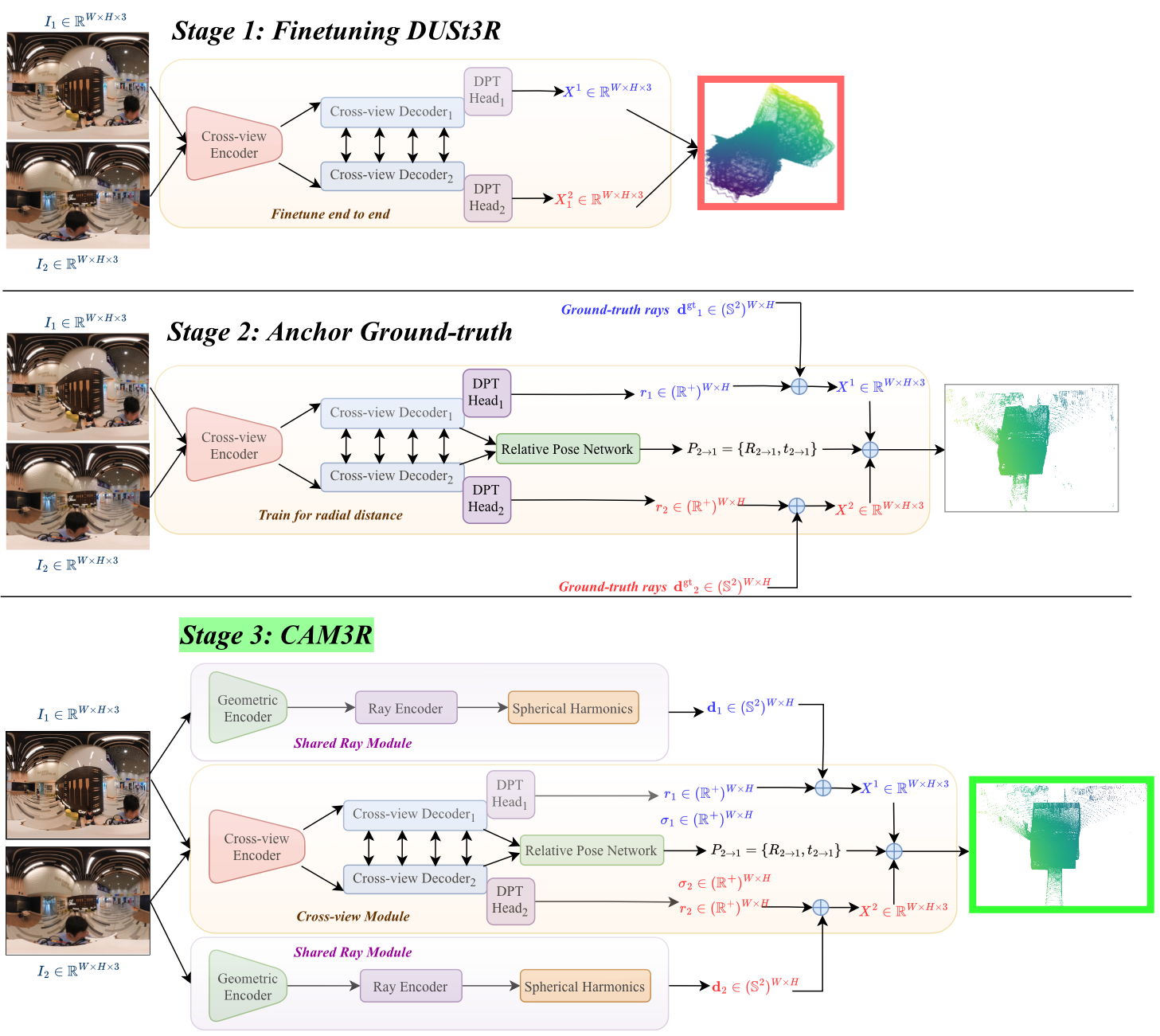}
    \caption{\textbf{Architectural evolution toward CAM3R.} Three stages explored during development: (1) direct fine-tuning of a DUSt3R-style model on distorted data, (2) decoupled radial prediction with ground-truth camera rays which is not always feasible during inference, and (3) the final CAM3R design with Ray Module and Cross-View Module.}
    \label{fig:arch_evolution_diagram}
\end{figure*}

The main paper introduces CAM3R as a camera-agnostic framework for dense two-view 3D reconstruction that explicitly separates camera geometry from scene geometry. Here we briefly summarize the architectural evolution that motivated the final design as shown in \cref{fig:arch_evolution_diagram}.

Our final architecture emerged from a series of investigations into how existing transformer-based 3D reconstruction models, such as DUSt3R \cite{wang2024dust3r}, behave under severe camera distortions.

\textbf{Stage 1: Direct fine-tuning of DUSt3R on distorted datasets.}
Our first approach naively fine-tuned a DUSt3R-style model on our diverse dataset mixture \cref{sec:Dat}. Because this paradigm regresses the second view's geometry directly into the first view's coordinate frame, it forces the network to implicitly entangle two factors: (a) non-linear intrinsic projection; (b) scene geometry and relative pose. Empirically, the network struggled to disentangle these conflicting signals, yielding severely distorted pointmaps.

\textbf{Stage 2: Decoupled radial prediction with ground-truth rays.}
To stabilize the optimization, we modified the attached Dense Prediction Transformer (DPT) \cite{ranftl2021vision} head to regress radial distances strictly in each view's independent local coordinate frame. We anchored the geometry using ground-truth camera rays which significantly stabilized the pointmap regression quality but relied on ground-truth camera intrinsics during inference.

\textbf{Stage 3: Learned ray estimation via a dedicated camera branch.}
To remove the dependency on ground-truth rays, we introduced a dedicated Ray Module, leveraging the weights of the angular module of UniK3D \cite{piccinelli2025unik3d}, to dynamically predict per-pixel ray directions. To align the independent local pointmaps into a shared 3D space, we appended a Relative Pose Network. This final formulation successfully separates camera calibration from scene reconstruction yielding the final CAM3R architecture.

\begin{figure*}[ht]
    \centering
    \includegraphics[width=\textwidth]{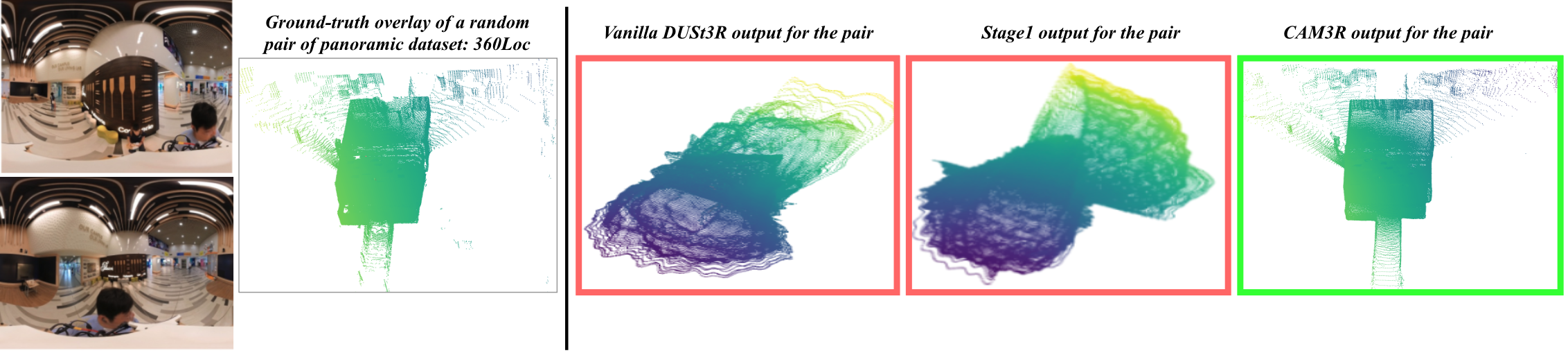}
    \caption{\textbf{Qualitative progression of architectural design choices.} From left to right: (1) Input image pairs (panoramas) with overlaid ground-truth 3D point clouds; (2) Reconstruction from \textbf{vanilla DUSt3R} \cite{wang2024dust3r}, which assumes pinhole projection; (3) Reconstruction from \textbf{fine-tuned DUSt3R (Stage 1)}, directly trained on distorted dataset mixture; and (4) Our final decoupled formulation, \textbf{CAM3R (Stage 3)}. While vanilla DUSt3R collapses under non-rectilinear distortions, naive fine-tuning fails to resolve the underlying geometric conflicts, forcing the network to generate distorted pointmaps. By explicitly decoupling internal ray geometry from relative depth, CAM3R recovers rigid, structurally accurate 3D geometry.}
    \label{fig:evolution_pipeline}
\end{figure*}

\begin{table*}[ht]
\centering
\caption{\textbf{Quantitative two-view validation of architectural design choices.} We report two-view relative pose estimation metrics at $15^\circ$ ($\uparrow$). Comparing Vanilla DUSt3R, a naively fine-tuned DUSt3R (Stage 1), and our final decoupled CAM3R (Stage 3) reveals that simply exposing standard foundation models to distorted data is mathematically insufficient. Our model yields massive performance gains, particularly on extreme optical manifolds (\eg 2D3DS \cite{armeni2017joint}  and 360Loc \cite{huang2024360loc}).}
\label{tab:evolution_metrics}
\resizebox{\textwidth}{!}{
\begin{tabular}{@{}lcccccccccc@{}}
\toprule
\textbf{Model Iteration} & \multicolumn{2}{c}{\textbf{2D3DS}} & \multicolumn{2}{c}{\textbf{MegaDepth}} & \multicolumn{2}{c}{\textbf{CO3Dv2 (Zero-shot)}} & \multicolumn{2}{c}{\textbf{360Loc}} & \multicolumn{2}{c}{\textbf{ADT}} \\
\cmidrule(lr){2-3} \cmidrule(lr){4-5} \cmidrule(lr){6-7} \cmidrule(lr){8-9} \cmidrule(lr){10-11}
& RRA@15 & RTA@15 & RRA@15 & RTA@15 & RRA@15 & RTA@15 & RRA@15 & RTA@15 & RRA@15 & RTA@15 \\
\midrule
Vanilla DUSt3R  \cite{wang2024dust3r} & 10.6 & 6.0 & 95.6 & 80.8 & 94.7 & 43.1 & 0.0 & 0.0 & 91.0 & 63.6 \\
Fine-tuned DUSt3R (Stage 1) & 17.8 & 10.9 & 94.2 & 72.5 & 94.3 & 52.7 & 13.0 & 9.2 & 87.5 & 65.4 \\
\midrule
\textbf{CAM3R (Final - Stage 3)} & \textbf{97.7} & \textbf{94.3} & \textbf{96.8} & \textbf{94.2} & \textbf{97.5} & \textbf{88.2} & \textbf{96.0} & \textbf{91.0} & \textbf{99.0} & \textbf{95.0} \\
\bottomrule
\end{tabular}}
\end{table*}

Quantitative and qualitative comparisons illustrating the progression from Stage~1 to the final model are provided in \cref{fig:evolution_pipeline,tab:evolution_metrics}.

\section{Training Configuration}
\label{subsec:training_config}

CAM3R is optimized end-to-end using the AdamW \cite{loshchilov2019decoupledweightdecayregularization} optimizer. We employ a linear warmup phase during the first few epochs to prevent early gradient instability, followed by a half-cycle cosine learning-rate decay. 
The complete training hyperparameters and scaling factors are summarized in \cref{tab:hyperparams_training}. 

\begin{table}[t]
\centering
\small
\renewcommand{\arraystretch}{1.15}
\caption{Optimization hyperparameters for end-to-end CAM3R training.}
\begin{tabular}{ll}
\toprule
\textbf{Hyperparameter} & \textbf{Value / Description} \\
\midrule
Batch Size (per GPU) & 4 \\
Gradient Accumulation & 2 steps \\
Effective Batch Size & 32 ($4$ GPUs $\times$ $4$ batch $\times$ $2$ accumulation) \\
Optimizer & AdamW \cite{loshchilov2019decoupledweightdecayregularization} ($\beta_1=0.9$, $\beta_2=0.95$) \\
Base Learning Rate (\texttt{blr}) & $1.5\times10^{-4}$ \\
Learning Rate Scaling & $\texttt{lr} = \texttt{blr} \times \frac{\texttt{effective batch size}}{256}$ \\
Minimum LR (\texttt{min\_lr}) & $1\times10^{-6}$ (cosine floor) \\
Learning Rate Schedule & Linear warmup (10 epochs) + cosine decay \\
Weight Decay & $0.05$ \\
Zero Weight Decay & Bias and LayerNorm parameters \\
Total Epochs & 300--500 \\
\bottomrule
\end{tabular}
\label{tab:hyperparams_training}
\end{table}

We detail the progressive spatial and channel dimensions of tensors as they propagate through the CAM3R pipeline in \cref{tab:cam3r_architecture}. 

\begin{table}[ht]
\centering
\caption{Architecture overview of CAM3R. $N_P = HW/P^2$ denotes the token sequence length for patch size $P$.}
\small
\resizebox{\textwidth}{!}{
\begin{tabular}{@{}llcc@{}}
\toprule
\textbf{Stage} & \textbf{Configuration} & \textbf{Input Shape} & \textbf{Output Shape} \\
\midrule
\multicolumn{4}{c}{\textbf{Ray Module for Camera Geometry}} \\
\midrule
Pixel Encoder & ViT-L/14 (24 layers), outputs $\{6,12,18,24\}$ & $B \times 3 \times H \times W$ & $\{B \times N_{14} \times 1024\}_{\times4}$ \\
Token Projection & Linear projection ($1024 \rightarrow 512$) & $\{B \times N_{14} \times 1024\}_{\times4}$ & $\{B \times N_{14} \times 512\}_{\times4}$ \\
Angular Regressor & 2 Transformer blocks, 8 heads, $D=512$ & $B \times 18 \times 512$ & $B \times 10$ (Intrinsics + SH coeffs.) \\
Ray Generation & FoV projection + SH basis (deg $\le3$) & Intrinsics + SH + pixel grid & $B \times H \times W \times 3$ \\
\midrule
\multicolumn{4}{c}{\textbf{Cross-View Module for Scene Geometry}} \\
\midrule
Patch Embedding & Linear projection, patch $16 \times 16$ & $B \times 3 \times H \times W$ & $B \times N_{16} \times 1024$ \\
Transformer Encoder & 24 blocks, 16 heads, $D=1024$ & $B \times N_{16} \times 1024$ & $B \times N_{16} \times 1024$ \\
Decoder Projection & Linear ($1024 \rightarrow 768$) & $B \times N_{16} \times 1024$ & $B \times N_{16} \times 768$ \\
Asymmetric Decoder & 12 blocks (self + cross attention) & $B \times N_{16} \times 768$ & $\{B \times N_{16} \times 768\}_{\times12}$ \\
DPT Head & Multi-scale fusion $\{0,6,9,12\}$, $D_{feat}=256$ & $\{B \times N_{16} \times 768\}_{\times4}$ & $B \times 2 \times H \times W$ \\
3D Point Computation & $P = r \cdot d$ (ray $\times$ distance) & Rays \& distance map & $B \times H \times W \times 3$ \\
Relative Pose Head & Global pooling + MLP ($D=256$) + SVD & Decoder features & $B \times 4 \times 4 \in SE(3)$ \\
\bottomrule
\end{tabular}}
\label{tab:cam3r_architecture}
\end{table}

\section{Ray-aware Global Alignment}
\label{sec:RGAsupp}

The main paper introduces the Ray-Aware Global Alignment framework for multi-view reconstruction. Here we provide additional implementation details regarding graph pruning, initialization, and optimization dynamics that improve global pose estimation.

\subsection{Multi-stage Scene-Graph Pruning}

Before global optimization, we apply a pruning cascade to remove unreliable edges from the pairwise scene graph. For each image pair $(i,j)$ the graph contains two directed edges $e_{ij}$ and $e_{ji}$ corresponding to the relative poses predicted by CAM3R.

\textbf{1. Symmetric Pose Consistency.} 
In a general unconstrained multi-view settings, neural pose regressors can occasionally hallucinate confident but erroneous relative poses, for instance, when confronted with wide-baseline pairs with minimal visual overlap. To efficiently filter these outliers before performing expensive alignment, we enforce a strict forward-backward cycle consistency as a heuristic \cite{zach2010disambiguating}.

For a valid prediction, the relative poses $\mathbf{P}_{ij}=\{\mathbf{R}_{ij},\mathbf{t}_{ij}\}$ and $\mathbf{P}_{ji}=\{\mathbf{R}_{ji},\mathbf{t}_{ji}\}$ must satisfy the inverse relation $\mathbf{R}_{ji} = \mathbf{R}_{ij}^\top$ and $\mathbf{t}_{ji} = -\mathbf{R}_{ij}^\top \mathbf{t}_{ij}$.

We measure the deviation using angular residuals
\begin{equation}
\Delta\mathbf{R}=\mathbf{R}_{ji}\mathbf{R}_{ij}, \qquad
\theta_{\text{rot}}=\arccos\!\left(\frac{\mathrm{tr}(\Delta\mathbf{R})-1}{2}\right)\frac{180^\circ}{\pi}
\end{equation}
\begin{equation}
\mathbf{t}_{\text{exp}}=-\mathbf{R}_{ij}^{\top}\mathbf{t}_{ij}, \qquad
\theta_{\text{tra}}=
\arccos\!\left(
\frac{\mathbf{t}_{ji}\cdot\mathbf{t}_{\text{exp}}}
{\|\mathbf{t}_{ji}\|_2\|\mathbf{t}_{\text{exp}}\|_2}
\right)\frac{180^\circ}{\pi}.
\end{equation}

Edges where $\theta_{\text{rot}}>\tau_{rot}$ or $\theta_{\text{tra}}>\tau_{tra}$  are discarded.

\textbf{2. Geometric Overlap via Mutual Nearest Neighbors (MNN).}

To verify dense geometric consistency, we transform the pointmap of view $j$ into the coordinate frame of view $i$ using $\mathbf{P}_{ij}$. KD-trees are built on both point sets and mutual nearest neighbors (MNN) are extracted. Let $n_e$ denote the number of mutual matches. Rather than using a fixed threshold, we apply an adaptive quantile gate: edges whose $n_e$ falls below the $20^{\text{th}}$ percentile of the scene-wide match distribution are removed. We additionally enforce strict symmetry: if $e_{ij}$ passes but $e_{ji}$ fails, both edges are discarded. This prevents asymmetric outliers from corrupting the graph.

\textbf{3. Largest Connected Component Extraction.} 

After pruning, the scene graph may fragment. We therefore retain only the largest component, ensuring that global optimization is performed on a single consistent camera graph.

\subsection{Anchor-Scale Initialization}
\begin{equation}
\min_{P_i, s_i}
\sum_{(i,j) \in E_{\text{pruned}}}
\sum_{\mathbf{u}}
\sigma_{i,j}(\mathbf{u})
\left\|
P_i \big( s_i \mathbf{x}_i(\mathbf{u}) \big)
-
P_j \big( s_j \mathbf{x}_j(\mathbf{u}) \big)
\right\|_2^2
\label{eq:global_opt}
\end{equation}

The global alignment objective \cref{eq:global_opt} (from the main paper) is sensitive to initialization. We compute a deterministic initialization before optimization. An anchor image $a^\ast$ is selected as the node with the highest degree in the pruned graph. Its pose is fixed as the global reference frame $\mathbf{T}_{a^\ast} = \mathbf{I}_{4\times4}$. Because pairwise reconstructions may exist in arbitrary scales, this initialization normalizes the global coordinate frame before optimization begins.

\subsection{Ray-Conditioned Optimization Dynamics}
With the above initialization, global alignment is performed using the objective (\cref{eq:global_opt}) described in the main paper. The optimization variables include camera poses $\mathbf{T}_i$, per-image scales $\log s_i$, and log-depth values $\log d_i$, while the unit ray fields $\mathbf{r}_i$ predicted by the Ray Module remain fixed. The effective 3D point during optimization is therefore
\begin{equation}
\mathbf{p}_i
=
d_{\text{eff},i}\mathbf{r}_i
=
\exp(\log d_i+\log s_i)\mathbf{r}_i.
\end{equation}

Freezing the ray directions is critical. Standard point cloud alignment \cite{wang2024dust3r} optimizes raw $(X,Y,Z)$ coordinates with a pinhole assumption, which can implicitly distort the camera projection geometry. By constraining points to move only along their optical rays $\mathbf{r}_i$, the optimizer preserves the camera-agnostic ray manifold while refining depth, pose, and scale.

Optimization is performed using AdamW \cite{loshchilov2019decoupledweightdecayregularization} with a cosine learning rate schedule. Pixel-wise confidence predictions $\sigma_i$ are used to down-weight unreliable geometric observations.

\section{Datasets}
\label{sec:Dat}
To learn a camera-agnostic geometric representation, our data corpus covers the fundamental projection models and scene types. We assemble a heterogeneous dataset mixture comprising \textbf{MegaDepth} \cite{li2018megadepth}, \textbf{360Loc} \cite{huang2024360loc}, \textbf{2D3DS} \cite{armeni2017joint}, and \textbf{Aria Digital Twin (ADT)} \cite{pan2023aria}. \textbf{MegaDepth} contains large-scale Internet imagery captured primarily with pinhole cameras and includes both indoor and outdoor scenes with wide viewpoint variation. \textbf{360Loc} provides panoramic imagery collected in expansive indoor and outdoor environments. \textbf{2D3DS} contains indoor panoramic captures of highly complex, cluttered building interiors. \textbf{ADT} provides egocentric fisheye imagery captured with wearable devices in indoor environments. 
The exact training and testing splits are summarized in \cref{tab:datasets}.

\begin{table}[ht]
\centering
\caption{Summary of dataset splits used for training and evaluation. We adhere to official protocols to ensure fair benchmarking across all projection domains.}
\label{tab:datasets}
\begin{tabular}{@{}lll@{}}
\toprule
\textbf{Dataset} & \textbf{Training Split} & \textbf{Testing Split} \\
\midrule
2D3DS \cite{armeni2017joint} & Areas 1, 2, 3, 4 & Areas 5a, 5b, 6 \\
ADT \cite{pan2023aria} & Apartment: $\sim$multi-user & Apartment: multi-user, Lite \\
MegaDepth \cite{li2018megadepth} & 0000--4541 (excl. 0015, 0022) & 0015, 0022, 5000--5018 \\
360Loc \cite{huang2024360loc} & Hall, Piatrium & Atrium, Concourse \\
\bottomrule
\vspace{-20pt}
\end{tabular}
\end{table}

\subsection{Training Pair Curation Strategy}
\label{subsec:training_pair}
For datasets lacking explicitly defined ground-truth image pairs, we curate training pairs using ground-truth poses with sufficient spatial overlap and meaningful wide-baseline. To support the two-phase training curriculum, \textbf{CAM3R-homo} and \textbf{CAM3R}, we construct \textit{homogeneous} (intra-projection) and \textit{heterogeneous} (cross-projection) image pairs, respectively. 

\textbf{2D3DS \cite{armeni2017joint}.} 
\textit{Homogeneous (Panorama-Panorama)} pairs are formed based on spatial separation. Let $\mathbf{c}_i, \mathbf{c}_j \in \mathbb{R}^3$ denote camera positions in world coordinates. We enforce a strict baseline interval of $d_{\min} \le \|\mathbf{c}_i - \mathbf{c}_j\|_2 \le d_{\max}$ where $d_{\min} = \text{0.1 m}$ and $d_{\max} = \text{2.2 m}$. These thresholds reflect the dense, cluttered nature of 2D3DS indoor environments. We retain the top-$K$ closest valid neighbors per anchor (default $K = 5$).

\textit{Heterogeneous (Panorama-Pinhole/Panorama-Fisheye)} are created from valid homogeneous neighbors where pinhole frames at the neighboring panorama's location are selected as targets. Frames with the highest overlap are selected, and synthetic fisheye versions are generated via equidistant projection.

\textbf{360Loc \cite{huang2024360loc}.}
\textit{Homogeneous} and \textit{Heterogeneous} pairs are created using the same logic as in 2D3DS.  In contrast to the tight rooms of 2D3DS, 360Loc features expansive outdoor areas and massive indoor atria. Hence, we expand the baseline threshold to $d_{\min}=\text{1.5 m,}\ d_{\max}=\text{10.0 m}$.

\textbf{ADT\cite{pan2023aria} .}
\textit{Homogeneous (Fisheye-Fisheye)} pairs are created based on the
baseline distance and relative viewing angle. Because ADT consists of egocentric head-mounted sequences, these thresholds are tailored to human motion dynamics. A pair is considered geometrically valid if $\text{0.35m} \le b \le \text{1.75m}$ and $25^\circ \le \theta \le 65^\circ$.

\textit{Heterogeneous (Fisheye-Pinhole)} pairs are created by extracting simultaneous fisheye and pinhole renders per frame using the Aria project toolkit, allowing us to form cross-modality pairs.

\textbf{MegaDepth \cite{li2018megadepth}.}
The pairs are provided directly from precomputed SfM and MVS metadata \cite{wang2024dust3r}. All pairs remain Pinhole-Pinhole, and no synthetic heterogeneity is injected during training.

\textbf{CO3Dv2\cite{reizenstein2021common}.}
To test zero-shot robustness, even-indexed frames remain pinhole while odd-indexed frames are synthetically warped to fisheye, rendering the sequences inherently heterogeneous.

\subsection{Evaluation Protocols}
\label{sec:eval_protocols}
\textbf{Two-View Evaluation.}
For pairwise evaluation, we assess the model's ability to reconstruct geometry from image pairs. We apply the exact same homogeneous and heterogeneous pairing logic, baseline thresholds, and overlap constraints described in \cref{sec:Dat}, but restrict the sampling strictly to the \textbf{Testing Splits} in \cref{tab:datasets}. To evaluate \textbf{zero-shot generalization}, we additionally utilize the \textbf{CO3Dv2} \cite{reizenstein2021common} dataset.

\textbf{Multi-View Evaluation.}
For multi-view evaluation, we assess the model's ability to achieve global consistency across entire test scenes in \cref{tab:datasets}. The evaluation graphs are constructed as follows to mirror realistic scenarios:
\begin{itemize}
    \item \textbf{MegaDepth.\cite{li2018megadepth}} In every test scene, evaluation groups are formed around \textbf{hub} images identified as having the highest number of co-visible connections in the SfM graph. The immediate first-order neighborhood (the first circle of matches) around each hub defines the multi-view evaluation set.
    \item \textbf{2D3DS \cite{armeni2017joint} \& 360Loc \cite{huang2024360loc}.} Multi-view evaluation set comprises all heterogenous images in the subscene/scene belonging to the test split.
    \item \textbf{ADT \cite{pan2023aria} \& CO3Dv2 \cite{reizenstein2021common}.} We load the video sequences in the test split with alternate Pinhole-Fisheye frames.
\end{itemize}

\subsection{Ground-Truth Rays and Pointmap Construction}
\textbf{Dense Pointmaps.}
For each view in  dataset $i$, the ground-truth 3D pointmap $\mathbf{\overline{X}}^{i,j}(\mathbf{u})$ in the camera coordinate frame is explicitly factorized as unit ray direction $\mathbf{\overline{d}}_i(\mathbf{u})$ and radial distance along the ray $\overline{r}_i(\mathbf{u})$. Only valid pixels ($\overline{r}_i > 0$) are included in the mask $\mathcal{V}(\mathbf{u})$ for training.

\subsubsection{Projection-Specific Ray Computation. }
\begin{itemize}
    \item \textbf{Equirectangular Panoramas.} Rays are computed deterministically from the pixel coordinates. Using the normalized pixel center convention $(u, v) = ((u_{\text{idx}}+0.5)/W, (v_{\text{idx}}+0.5)/H)$, we map to longitude $\lambda = (u - 0.5)\, 2\pi$ and latitude $\phi= (0.5 - v)\, \pi$.

    \item \textbf{Pinhole Cameras.} Rays are derived from the intrinsics matrix $\mathbf{K}$ via inverse projection. Perspective datasets typically provide Z-depth, $D_z(\mathbf{u})$, which measures the depth strictly along the optical axis. To maintain spatial consistency with our spherical formulation, we convert this Z-depth to radial distance.

    \item \textbf{Fisheye Lenses.} Rays are computed per pixel using the exact sensor calibration unprojection routines, which account for radial distortions. 
 
\end{itemize}

Once the local pointmaps $\mathbf{\overline{X}}^{i,i}(\mathbf{u})$ are constructed, they can be expressed in the global world frame using the camera-to-world pose $\mathbf{\overline{P}}_i = (\mathbf{\overline{R}}_i, \mathbf{\overline{t}}_i) \in SE(3)$. 
During training, we project view $i$'s pointmap into the coordinate frame of view $j$ to establish relative geometric supervision. 

\begin{equation}
\begin{aligned}
\mathbf{\overline{X}}^{\text{world}}(\mathbf{u})
&=
\mathbf{\overline{R}}_i \mathbf{\overline{X}}^{i,i}(\mathbf{u}) + \mathbf{\overline{t}}_i,
\qquad
\mathbf{\overline{X}}^{i,j}(\mathbf{u})
=
\mathbf{\overline{R}}_{j \leftarrow i} \mathbf{\overline{X}}^{i,i}(\mathbf{u})
+
\mathbf{\overline{t}}_{j \leftarrow i}.
\end{aligned}
\end{equation}
\[
\mathbf{\overline{R}}_{j \leftarrow i} = \mathbf{\overline{R}}_j \mathbf{\overline{R}}_i^\top,
\qquad
\mathbf{\overline{t}}_{j \leftarrow i} = \mathbf{\overline{t}}_j - \mathbf{\overline{R}}_{j \leftarrow i} \mathbf{\overline{t}}_i .
\]

\section{Limitations and Future Directions}
\label{sec:limitations}
CAM3R introduces an explicit decoupling between camera geometry and scene reconstruction, which improves robustness across heterogeneous projection models but introduces certain limitations.

\textbf{Architectural Overhead.}
The current design employs two separate Vision Transformer (ViT) \cite{dosovitskiy2021imageworth16x16words} backbones for the Ray Module and Cross-View Module, resulting in higher memory usage. While the decoupled learning is important for strong distortions, future work could reduce redundancy through parameter sharing or knowledge distillation to learn a unified encoder \cite{sariyildiz2025dune}.

\textbf{Pairwise Scalability.}
CAM3R runs for $O(N^2)$ pairwise inferences to construct the scene graph for multi-view reconstruction. Although our pruning strategy mitigates the cost before global alignment, scaling to very large image collections remains expensive. Extending the ray-based formulation to multi-view transformer architectures \cite{wang2025vggt, wang2025pi} is a promising direction for improving scalability.

Despite these limitations, our results suggest that explicitly modeling camera rays provides a foundation for camera-agnostic 3D reconstruction.

\bibliographystyle{splncs04}
\bibliography{main}

@String(CVPR  = {IEEE Conf. Comput. Vis. Pattern Recog.})

@String(ECCV  = {Eur. Conf. Comput. Vis.})

@String(CVPR  = {CVPR})

@String(ECCV  = {ECCV})

@inproceedings{wang2024dust3r,
  title={Dust3r: Geometric 3d vision made easy},
  author={Wang, Shuzhe and Leroy, Vincent and Cabon, Yohann and Chidlovskii, Boris and Revaud, Jerome},
  booktitle={Proceedings of the IEEE/CVF conference on computer vision and pattern recognition},
  pages={20697--20709},
  year={2024}
}

@inproceedings{leroy2024grounding,
  title={Grounding image matching in 3d with mast3r},
  author={Leroy, Vincent and Cabon, Yohann and Revaud, J{\'e}r{\^o}me},
  booktitle={European conference on computer vision},
  pages={71--91},
  year={2024},
  organization={Springer}
}

@inproceedings{jang2025pow3r,
  title={Pow3r: Empowering unconstrained 3d reconstruction with camera and scene priors},
  author={Jang, Wonbong and Weinzaepfel, Philippe and Leroy, Vincent and Agapito, Lourdes and Revaud, Jerome},
  booktitle={Proceedings of the Computer Vision and Pattern Recognition Conference},
  pages={1071--1081},
  year={2025}
}

@inproceedings{wang2025vggt,
  title={Vggt: Visual geometry grounded transformer},
  author={Wang, Jianyuan and Chen, Minghao and Karaev, Nikita and Vedaldi, Andrea and Rupprecht, Christian and Novotny, David},
  booktitle={Proceedings of the Computer Vision and Pattern Recognition Conference},
  pages={5294--5306},
  year={2025}
}

@article{wang2025pi,
  title = {{$\pi^3$: Permutation-Equivariant Visual Geometry Learning}},
  author = {Wang, Yifan and Zhou, Jianjun and Zhu, Haoyi and Chang, Wenzheng and Zhou, Yang and Li, Zizun and Chen, Junyi and Pang, Jiangmiao and Shen, Chunhua and He, Tong},
  journal = {arXiv preprint arXiv:2507.13347},
  year = {2025}
}

@inproceedings{schoenberger2016sfm,
    author={Sch\"{o}nberger, Johannes Lutz and Frahm, Jan-Michael},
    title={{Structure-from-Motion Revisited}},
    booktitle={Conference on Computer Vision and Pattern Recognition (CVPR)},
    year={2016}
}

@inproceedings{reizenstein2021common,
  title={Common objects in 3d: Large-scale learning and evaluation of real-life 3d category reconstruction},
  author={Reizenstein, Jeremy and Shapovalov, Roman and Henzler, Philipp and Sbordone, Luca and Labatut, Patrick and Novotny, David},
  booktitle={Proceedings of the IEEE/CVF international conference on computer vision},
  pages={10901--10911},
  year={2021}
}

@inproceedings{li2018megadepth,
  title={Megadepth: Learning single-view depth prediction from internet photos},
  author={Li, Zhengqi and Snavely, Noah},
  booktitle={Proceedings of the IEEE conference on computer vision and pattern recognition},
  pages={2041--2050},
  year={2018}
}

@inproceedings{piccinelli2025unik3d,
  title={Unik3d: Universal camera monocular 3d estimation},
  author={Piccinelli, Luigi and Sakaridis, Christos and Segu, Mattia and Yang, Yung-Hsu and Li, Siyuan and Abbeloos, Wim and Van Gool, Luc},
  booktitle={Proceedings of the Computer Vision and Pattern Recognition Conference},
  pages={1028--1039},
  year={2025}
}

@article{kannala2006generic,
  title={A generic camera model and calibration method for conventional, wide-angle, and fish-eye lenses},
  author={Kannala, Juho and Brandt, Sami S},
  journal={IEEE transactions on pattern analysis and machine intelligence},
  volume={28},
  number={8},
  pages={1335--1340},
  year={2006},
  publisher={IEEE}
}

@inproceedings{geyer2000unifying,
  title={A unifying theory for central panoramic systems and practical implications},
  author={Geyer, Christopher and Daniilidis, Kostas},
  booktitle={European conference on computer vision},
  pages={445--461},
  year={2000},
  organization={Springer}
}

@inproceedings{detone2018superpoint,
  title={Superpoint: Self-supervised interest point detection and description},
  author={DeTone, Daniel and Malisiewicz, Tomasz and Rabinovich, Andrew},
  booktitle={Proceedings of the IEEE conference on computer vision and pattern recognition workshops},
  pages={224--236},
  year={2018}
}

@inproceedings{sarlin2020superglue,
  title={Superglue: Learning feature matching with graph neural networks},
  author={Sarlin, Paul-Edouard and DeTone, Daniel and Malisiewicz, Tomasz and Rabinovich, Andrew},
  booktitle={Proceedings of the IEEE/CVF conference on computer vision and pattern recognition},
  pages={4938--4947},
  year={2020}
}

@inproceedings{edstedt2024roma,
  title={Roma: Robust dense feature matching},
  author={Edstedt, Johan and Sun, Qiyu and B{\"o}kman, Georg and Wadenb{\"a}ck, M{\aa}rten and Felsberg, Michael},
  booktitle={Proceedings of the IEEE/CVF conference on computer vision and pattern recognition},
  pages={19790--19800},
  year={2024}
}

@article{lowe2004distinctive,
  title={Distinctive image features from scale-invariant keypoints},
  author={Lowe, David G},
  journal={International journal of computer vision},
  volume={60},
  number={2},
  pages={91--110},
  year={2004},
  publisher={Springer}
}

@inproceedings{schoenberger2016mvs,
    author={Sch\"{o}nberger, Johannes Lutz and Zheng, Enliang and Pollefeys, Marc and Frahm, Jan-Michael},
    title={{Pixelwise View Selection for Unstructured Multi-View Stereo}},
    booktitle={European Conference on Computer Vision (ECCV)},
    year={2016}
}

@article{sturm2011camera,
  title={Camera models and fundamental concepts used in geometric computer vision},
  author={Sturm, Peter and Ramalingam, Srikumar and Tardif, Jean-Philippe and Gasparini, Simone and Barreto, Joao},
  journal={Foundations and Trends in Computer Graphics and Vision},
  volume={6},
  number={1-2},
  pages={1--183},
  year={2011},
  publisher={Emerald Publishing Limited}
}

@misc{liao2025deeplearningcameracalibration,
      title={Deep Learning for Camera Calibration and Beyond: A Survey}, 
      author={Kang Liao and Lang Nie and Shujuan Huang and Chunyu Lin and Jing Zhang and Yao Zhao and Moncef Gabbouj and Dacheng Tao},
      year={2025},
      eprint={2303.10559},
      archivePrefix={arXiv},
      primaryClass={cs.CV},
      url={https://arxiv.org/abs/2303.10559}, 
}

@article{vandenhende2021multi,
  title={Multi-task learning for dense prediction tasks: A survey},
  author={Vandenhende, Simon and Georgoulis, Stamatios and Van Gansbeke, Wouter and Proesmans, Marc and Dai, Dengxin and Van Gool, Luc},
  journal={IEEE transactions on pattern analysis and machine intelligence},
  volume={44},
  number={7},
  pages={3614--3633},
  year={2021},
  publisher={IEEE}
}

@article{yu2020gradient,
  title={Gradient surgery for multi-task learning},
  author={Yu, Tianhe and Kumar, Saurabh and Gupta, Abhishek and Levine, Sergey and Hausman, Karol and Finn, Chelsea},
  journal={Advances in neural information processing systems},
  volume={33},
  pages={5824--5836},
  year={2020}
}

@article{brown1971close,
  title={Close-range camera calibration},
  author={Brown, Duane C},
  journal={Photogrammetric engineering},
  volume={37},
  number={8},
  pages={855--866},
  year={1971}
}

@article{conrady1919lens,
  title={Lens-systems, decentered},
  author={Conrady, AE},
  journal={Monthly Notices of the Royal Astronomical Society, Vol. 79, p. 384-390},
  volume={79},
  pages={384--390},
  year={1919}
}

@article{urban2015improved,
  title={Improved wide-angle, fisheye and omnidirectional camera calibration},
  author={Urban, Steffen and Leitloff, Jens and Hinz, Stefan},
  journal={ISPRS Journal of Photogrammetry and Remote Sensing},
  volume={108},
  pages={72--79},
  year={2015},
  publisher={Elsevier}
}

@inproceedings{bogdan2018deepcalib,
  title={DeepCalib: A deep learning approach for automatic intrinsic calibration of wide field-of-view cameras},
  author={Bogdan, Oleksandr and Eckstein, Viktor and Rameau, Francois and Bazin, Jean-Charles},
  booktitle={Proceedings of the 15th ACM SIGGRAPH European Conference on Visual Media Production},
  pages={1--10},
  year={2018}
}

@inproceedings{veicht2024geocalib,
  title={Geocalib: Learning single-image calibration with geometric optimization},
  author={Veicht, Alexander and Sarlin, Paul-Edouard and Lindenberger, Philipp and Pollefeys, Marc},
  booktitle={European Conference on Computer Vision},
  pages={1--20},
  year={2024},
  organization={Springer}
}

@inproceedings{triggs1999bundle,
  title={Bundle adjustment—a modern synthesis},
  author={Triggs, Bill and McLauchlan, Philip F and Hartley, Richard I and Fitzgibbon, Andrew W},
  booktitle={International workshop on vision algorithms},
  pages={298--372},
  year={1999},
  organization={Springer}
}

@article{neal2011distributed,
  title={Distributed optimization and statistical learning via the alternating direction method of multipliers},
  author={Neal, Parikh and Eric, Chu and Borja, Peleato and Jonathan, Eckstein},
  journal={Foundations and Trends{\textregistered} in Machine learning},
  volume={3},
  number={1},
  pages={1--122},
  year={2011},
  publisher={Emerald Publishing Limited}
}

@article{armeni2017joint,
  title={Joint 2d-3d-semantic data for indoor scene understanding},
  author={Armeni, Iro and Sax, Sasha and Zamir, Amir R and Savarese, Silvio},
  journal={arXiv preprint arXiv:1702.01105},
  year={2017}
}

@inproceedings{huang2024360loc,
  title={360loc: A dataset and benchmark for omnidirectional visual localization with cross-device queries},
  author={Huang, Huajian and Liu, Changkun and Zhu, Yipeng and Cheng, Hui and Braud, Tristan and Yeung, Sai-Kit},
  booktitle={Proceedings of the IEEE/CVF conference on computer vision and pattern recognition},
  pages={22314--22324},
  year={2024}
}

@inproceedings{pan2023aria,
  title={Aria digital twin: A new benchmark dataset for egocentric 3d machine perception},
  author={Pan, Xiaqing and Charron, Nicholas and Yang, Yongqian and Peters, Scott and Whelan, Thomas and Kong, Chen and Parkhi, Omkar and Newcombe, Richard and Ren, Yuheng Carl},
  booktitle={Proceedings of the IEEE/CVF International Conference on Computer Vision},
  pages={20133--20143},
  year={2023}
}

@misc{dosovitskiy2021imageworth16x16words,
      title={An Image is Worth 16x16 Words: Transformers for Image Recognition at Scale}, 
      author={Alexey Dosovitskiy and Lucas Beyer and Alexander Kolesnikov and Dirk Weissenborn and Xiaohua Zhai and Thomas Unterthiner and Mostafa Dehghani and Matthias Minderer and Georg Heigold and Sylvain Gelly and Jakob Uszkoreit and Neil Houlsby},
      year={2021},
      eprint={2010.11929},
      archivePrefix={arXiv},
      primaryClass={cs.CV},
      url={https://arxiv.org/abs/2010.11929}, 
}

@inproceedings{ranftl2021vision,
  title={Vision transformers for dense prediction},
  author={Ranftl, Ren{\'e} and Bochkovskiy, Alexey and Koltun, Vladlen},
  booktitle={Proceedings of the IEEE/CVF international conference on computer vision},
  pages={12179--12188},
  year={2021}
}

@misc{loshchilov2019decoupledweightdecayregularization,
      title={Decoupled Weight Decay Regularization}, 
      author={Ilya Loshchilov and Frank Hutter},
      year={2019},
      eprint={1711.05101},
      archivePrefix={arXiv},
      primaryClass={cs.LG},
      url={https://arxiv.org/abs/1711.05101}, 
}

@article{kabsch1976solution,
  title={A solution for the best rotation to relate two sets of vectors},
  author={Kabsch, Wolfgang},
  journal={Foundations of Crystallography},
  volume={32},
  number={5},
  pages={922--923},
  year={1976},
  publisher={International Union of Crystallography}
}

@inproceedings{bengio2009curriculum,
  title={Curriculum learning},
  author={Bengio, Yoshua and Louradour, J{\'e}r{\^o}me and Collobert, Ronan and Weston, Jason},
  booktitle={Proceedings of the 26th annual international conference on machine learning},
  pages={41--48},
  year={2009}
}

@article{chang2017matterport3d,
  title={Matterport3d: Learning from rgb-d data in indoor environments},
  author={Chang, Angel and Dai, Angela and Funkhouser, Thomas and Halber, Maciej and Niessner, Matthias and Savva, Manolis and Song, Shuran and Zeng, Andy and Zhang, Yinda},
  journal={arXiv preprint arXiv:1709.06158},
  year={2017}
}

@inproceedings{gunes2025fiord,
  title={Fiord: A fisheye indoor-outdoor dataset with lidar ground truth for 3d scene reconstruction and benchmarking},
  author={Gunes, Ulas and Turkulainen, Matias and Ren, Xuqian and Solin, Arno and Kannala, Juho and Rahtu, Esa},
  booktitle={Scandinavian Conference on Image Analysis},
  pages={3--17},
  year={2025},
  organization={Springer}
}

@inproceedings{sariyildiz2025dune,
  title={Dune: Distilling a universal encoder from heterogeneous 2d and 3d teachers},
  author={Sar{\i}y{\i}ld{\i}z, Mert B{\"u}lent and Weinzaepfel, Philippe and Lucas, Thomas and De Jorge, Pau and Larlus, Diane and Kalantidis, Yannis},
  booktitle={Proceedings of the Computer Vision and Pattern Recognition Conference},
  pages={30084--30094},
  year={2025}
}

@inproceedings{zach2010disambiguating,
  title={Disambiguating visual relations using loop constraints},
  author={Zach, Christopher and Klopschitz, Manfred and Pollefeys, Marc},
  booktitle={2010 IEEE Computer Society Conference on Computer Vision and Pattern Recognition},
  pages={1426--1433},
  year={2010},
  organization={IEEE}
}

\end{document}